\documentclass[journal]{IEEEtran}
\ifCLASSINFOpdf
   \usepackage[pdftex]{graphicx}
\else
   \usepackage[dvips]{graphicx}
\fi
\usepackage{amsmath,amssymb}
\usepackage{algpseudocode}
\usepackage{algorithm}
\usepackage{times}
\usepackage{bbm}
\usepackage{epsfig}
\usepackage{graphicx}
\usepackage{times}
\usepackage{xcolor}
\usepackage{dsfont}
\usepackage{multirow}
\usepackage{url}
\usepackage{amsmath,amssymb}
\begin{document}
%
\title{Deep Mixture of Diverse Experts for Large-Scale Visual Recognition}
%
%
%

\author{Tianyi~Zhao, ~Jun~ Yu, ~Zhenzhong~Kuang, ~Wei~Zhang, ~Jianping~Fan
}

\maketitle

\begin{abstract}
In this paper, a deep mixture of diverse experts algorithm is developed for seamlessly combining a set of base deep CNNs (convolutional neural networks) with diverse outputs (task spaces), e.g., such base deep CNNs are trained to recognize different subsets of tens of thousands of atomic object classes. First, a two-layer (category layer and object class layer) ontology is constructed to achieve more effective solution for task group generation, e.g., assigning the semantically-related atomic object classes at the sibling leaf nodes into the same task group because they may share similar learning complexities. Second, one particular base deep CNNs with $M+1$ ($M \leq 1,000$) outputs is learned for each task group to recognize its $M$ atomic object classes effectively and identify one special class of ``not-in-group" automatically, and the network structure (numbers of layers and units in each layer) of the well-designed AlexNet is directly used to configure such base deep CNNs. A deep multi-task learning algorithm is developed to leverage the inter-class visual similarities to learn more discriminative base deep CNNs and multi-task softmax for enhancing the separability of the atomic object classes in the same task group. Finally, all these base deep CNNs with diverse outputs (task spaces) are seamlessly combined to form a deep mixture of diverse experts for recognizing tens of thousands of atomic object classes. Our experimental results have demonstrated that our deep mixture of diverse experts algorithm can achieve very competitive results on large-scale visual recognition.
\end{abstract}


%
\IEEEpeerreviewmaketitle

\section{Introduction}
%
%
%
%
\IEEEPARstart{L}arge-scale visual recognition (e.g., recognizing tens of thousands of atomic object classes) has recently received considerable attentions  [1-6], but it is still a challenging issue even the state-of-the-art methods have achieved impressive progresses. By learning high-level features and a $N$-way softmax in an end-to-end multi-layer manner, deep learning [11-18] has demonstrated its outstanding performance on recognizing large numbers of object classes because of its strong ability on learning highly invariant and discriminant features. Most successful designs of deep networks [11-17] optimize both their structures (i.e., numbers of layers and units in each layer) and their node weights for recognizing 1,000 object classes, and AlexNet [11-13] is the most popular design of deep CNNs (convolutional neural networks). Thus it is very nature for us to ask ourselves the following question: {\em How can we leverage some successful designs of the deep CNNs for 1,000 object classes to recognize tens of thousands of atomic object classes?} 

Because both the high-level features for image content representation and the $N$-way softmax for visual recognition are trained jointly in an end-to-end fashion, simply enlarging the final outputs of some well-designed deep CNNs (from $1,000$-way softmax into $N$-way one, $N \geq 10,000$) may not be an optimal solution, and the following three directions can be exploited.  

The {\bf first} direction is to re-shape the network structures by using more layers (depths) and more units in each layer (widths). In general, the deep CNNs can fully unfold their potentials when they are wider (more units per layer) and deeper (more layers) [14-17]. Some guidelines have been provided for designing the deep CNNs, but such guidelines have not provided theoretical solutions on how to set the optimal network structures (i.e., numbers of layers and units per layer). Identifying the optimal network structures has historically been relegated to manual optimization, which relies in human intuition and domain knowledge in conjunction with extensive trials and errors. 

When the image sets are sufficiently large and strong computation resources are also available, directly using more layers and more units (in each layer) to configure huge deep CNNs could be very intuitive and motivated for large-scale visual recognition application, but it may seriously suffer from the following problems: (a) {\em Unknown network structure}: It may require extensive trials and errors to identify the optimal network structure (i.e., numbers of layers and units in each layer) for such huge deep CNNs;  (b) {\em Huge training cost}: We may need large computation resources (that could be unavailable for most researchers) and spend several weeks or even several months to optimize such huge deep CNNs (i.e., both the network structures and the node weights); (c) {\em Overfitting}: Such huge deep CNNs may have hundreds of millions of node weights that may easily overfit the available training images (i.e., the number of trainable node weights could be much bigger than the number of available training images), and its performance may severely depend on careful tuning of hundreds of millions of node weights; (d) {\em Applicability}: High demand of computation resources may severely hinder deployment of such huge deep CNNs in many applications which have straight limitations and constraints on computation resources; (e) {\em Local optimum}: Large numbers of atomic object classes may have huge diversities on the inter-class visual similarities, thus the gradients of their objective function are heavily nonuniform and the underlying learning process may distract on discerning some object classes that are typically hard to be discriminated. 

The {\bf second} direction is to use transfer learning [53-61]. Unfortunately, most existing transfer learning approaches focus on transferring the deep networks (i.e., both the network structures and the node weights) learned for a big domain (with large task space) into another smaller domain (with small task space) when their task spaces are completely overlapped, e.g., the object classes in a smaller domain are just a subset of the object classes in a big domain. Some pioneer researches [62-63] have been done on leveraging the deep CNNs trained for recognizing 1,000 object classes in ImageNet1K to speed up the learning of the deep CNNs for recognizing 20 object classes in PASCAL VOC. Obviously, 20 object classes in PASCAL VOC [65-66] are just a small subset of 1,000 object classes in ImageNet1K, thus the deep networks learned for ImageNet1K (a big domain) can be transferred successfully for PASCAL VOC application (a smaller domain). On the other hand, our goal is to combine a set of base deep CNNs for 1,000 object classes (a small domain) to recognize tens of thousands of atomic object classes (a bigger domain). 

The {\bf third} direction is to combine a set of base deep CNNs. It is worth noting that all the existing techniques  focus on combining multiple deep CNNs which are trained to recognize the same set of object classes [39-52], e.g., all the base deep CNNs being combined share the same task spaces  (outputs). According to the best of our knowledge, there does not exist any approach for combining a set of base deep CNNs which have diverse outputs (task spaces), e.g., all these base deep CNNs are trained to recognize different subsets of tens of thousands of atomic object classes rather than the same set of atomic object classes. Without supporting effective combination of multiple base deep CNNs with diverse outputs (task spaces), we cannot leverage the well-designed deep CNNs for 1,000 object classes (such as AlexNet [11-13]) to recognize tens of thousands of atomic object classes.  Obviously, it is not straightforward to combine a set of base deep CNNs which are originally trained to recognize different subsets of tens of thousands of atomic object classes (i.e., their task spaces are different and diverse).

There are at least three challenges for combining a set of base deep CNNs with diverse outputs (task spaces): 

(1) {\bf\em Task Group Generation:} In order to leverage the network structure (i.e., number of layers and number of units in each layer) of the well-designed AlexNet for 1,000 object classes [11-13] to configure our base deep CNNs, large numbers of atomic object classes are first assigned into a set of task groups and each task group contains $M$ ($M \leq 1,000$) atomic object classes. If a random process is used for task group generation (i.e., random selection of $M$ ($M \leq 1,000$) atomic object classes for each task group), some atomic object classes with similar learning complexities could be assigned into different task groups, as a result, they may not be able to receive sufficient comparison and contrasting from each other.  

(2) {\bf\em Global Optimum:}  In most existing deep learning schemes, a $N$-way softmax is used and the inter-class correlations are completely ignored, as a result, the process for learning the deep CNNs may be pushed away from the global optimum because the gradients of the objective function are not uniform for all the object classes and such learning process may distract on discerning some object classes that are typically hard to be discriminated. 

(3) {\bf\em Comparability of Predictions:} For a given image or object proposal, all these base deep CNNs (for different task groups) could provide their individual predictions with certain scores, but the predictions from different base deep CNNs could be conflict and they are typically incomparable because these base deep CNNs are not trained jointly. 

Based on these observations, a deep mixture of diverse experts algorithm is developed in this paper for seamlessly combining a set of base deep CNNs with diverse outputs (task spaces), where the network structure of the well-designed AlexNet  [11-13] is directly used to configure such base deep CNNs and all these base deep CNNs are learned to recognize different subsets of tens of thousands of atomic object classes rather than the same set of atomic object classes. 

The rest of the paper is organized as: Section 2 briefly reviews the related work;  Section 3 introduces our algorithm for constructing a two-layer ontology; Section 4 presents our deep mixture of diverse experts algorithm for seamlessly combining a set of base deep CNNs with diverse outputs (task spaces); Section 5 reports our experimental results; and we conclude this paper at Section 6.

\section{Related Work} 
In this section, we briefly review the most relevant researches on: (1) deep learning [11-18]; (2) mixture of deep CNNs [39-52], (3) transfer learning [53-64], and (4) tree structures for hierarchical indexing of large numbers of object classes [23-35].

By learning high-level features and a $N$-way softmax jointly in an end-to-end multi-layer manner, deep learning [11-18] has demonstrated its outstanding performance on significantly boosting the accuracy rates on visual recognition.  Most successful designs of the deep CNNs optimize both their network structures (number of layers and number of units in each layer) and their node weights for recognizing 1,000 object classes, but simply enlarging the network structures (such as enlarging the widths and depths of the deep CNNs) to configure huge deep CNNs may require huge computation cost, and such huge deep CNNs may have hundreds of millions of node weights that may easily overfit the available training images. Obviously, the performance of such huge deep CNNs may largely depend on careful tuning of its hundreds of millions of node weights. On the other hand, simply increasing the softmax outputs (from $1,000$-way softmax into $N$-way one, $N \geq 10,000$) may not be able to achieve good results because the underlying deep CNNs (learned for 1,000 object classes) could be insufficient and inefficient to extract discriminative representations for tens of thousands of atomic object classes. 

By considering multiple inter-related learning tasks jointly, deep multi-task learning [19-22] has demonstrated its strong ability on learning more discriminative deep CNNs and multi-task softmax. Even deep multi-task learning has demonstrated many advantages in theory, one obstacle for applying deep multi-task learning to support large-scale visual recognition is how to identify the inter-related learning tasks automatically. One way to identify the inter-related learning tasks is to organize large numbers of object classes hierarchically in a tree structure [23, 34-35], and both the semantic ontology and the label tree or visual hierarchy have been explored [23-35]. 

To improve the accuracy rates on recognizing the same set of object classes, traditional deep mixture techniques aim to combine the predictions from multiple base deep CNNs when they are trained to recognize the same set of object classes (i.e., they share the same task space) [39-52]. In order to enhance the diversity of the base deep CNNs being combined, they are usually trained over different sample subsets, so that they may make their errors in different ways or even compensate each other. Ge et al. [39] have developed a mixture of deep CNNs (MixDCNN) by partitioning the training images into multiple subsets and learning one particular base deep CNNs for each image subset. In such MixDCNN approach, each of these base deep CNNs concentrates on learning the subtle differences for the same set of object classes, and all these base deep CNNs share the same task space (i.e., the same set of outputs). On the other hand, our deep mixture of diverse experts algorithm focuses on combining a set of base deep CNNs with diverse outputs (task spaces), e.g., all these base deep CNNs are trained to recognize different subsets of tens of thousands of atomic object classes rather than the same set of atomic object classes.

Transfer learning [53-63] has recently received enough attentions and it can be used to adapt the deep CNNs learned for a big domain (with large task space) into another smaller domain (with smaller task space), e.g., the object classes in a small domain are just a subset of the object classes in a big domain. Recently, some pioneer researches [62-63] have been done on leveraging the deep CNNs learned for recognizing 1,000 object classes in ImageNet1K to speed up the learning of the deep CNNs for recognizing 20 object classes in PASCAL VOC [65-66]. On the other hand, our deep mixture of diverse experts algorithm focuses on combining a set of base deep CNNs for 1,000 atomic object classes (a small domain) to recognize tens of thousands of atomic object classes (a big domain), e.g., from a small task space to a big one. Recently, Li and Hoiem [64] have developed an interesting approach, called learning without forgetting, to learn the deep networks incrementally when new object classes appear over time and the number of new object classes is properly very small as compared with the number of known object classes (whose deep networks are already learned).

\section{Two-Layer Ontology for Task Assignment}
ImageNet10K image set [24] is used in this paper for algorithm evaluation and it contains $10,184$ image categories, but not all of them are semantically atomic (mutually exclusive) because some of them are selected from high-level non-leaf nodes of the concept ontology. Thus the concept ontology used in ImageNet10K is incorporated to decompose such high-level image categories (from the non-leaf nodes of the concept ontology) into the most relevant atomic object classes (at the sibling leaf nodes of the concept ontology), and  $7,756$ atomic object classes are finally identified.  
\begin{figure}[t]
\vspace*{-0.188cm}
\begin{center}
\includegraphics[width=0.48\textwidth]{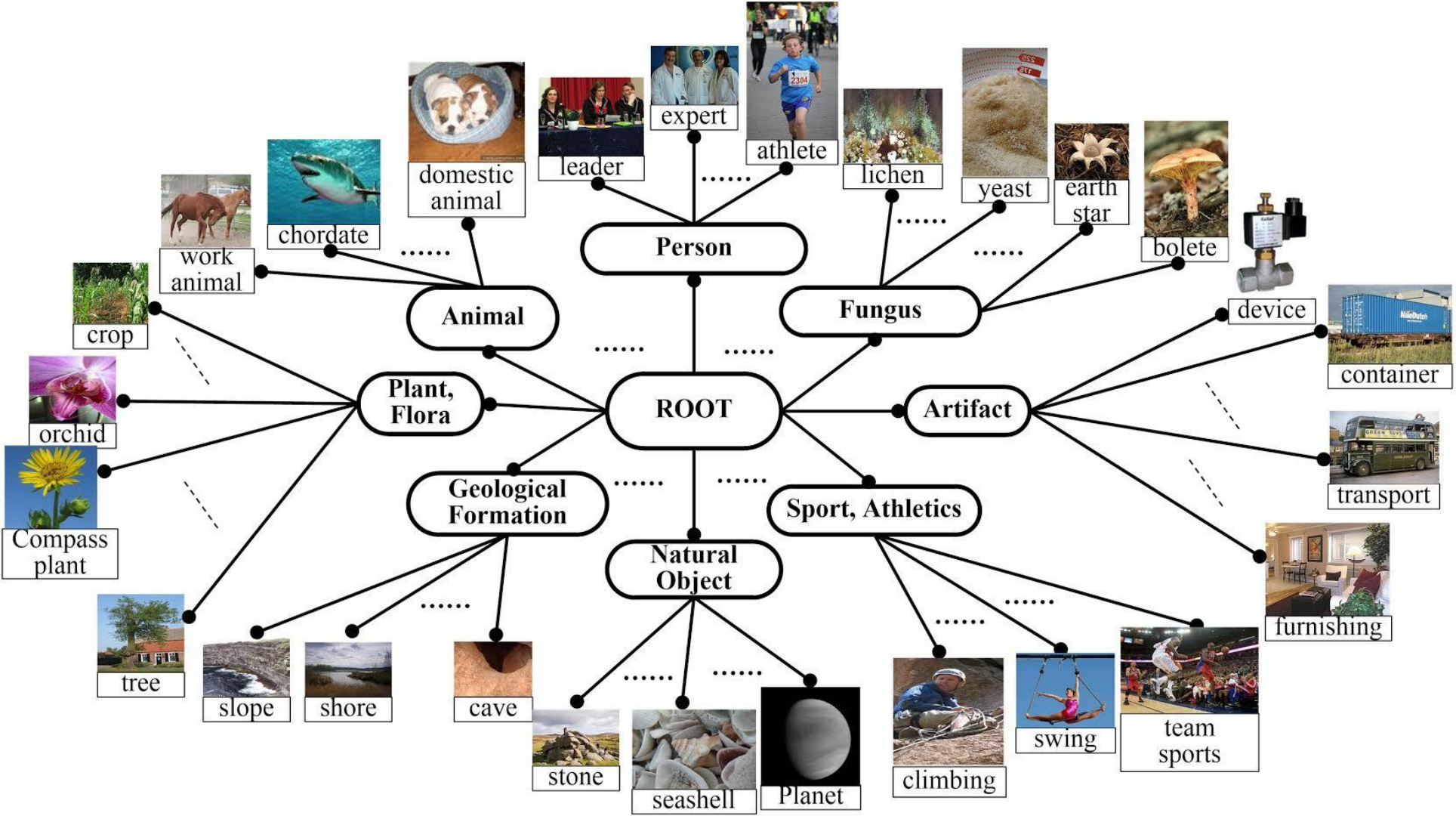}
\end{center}
\vspace*{-0.488cm}
   \caption{\bf Part of our two-layer ontology for indexing $7,756$ atomic object classes in ImageNet10K image set.}
\label{fig:long}
\label{fig:onecol}
\vspace*{-0.118cm}
\end{figure}
\begin{figure}[t]
\vspace*{-0.188cm}
\begin{center}
\includegraphics[width=0.48\textwidth]{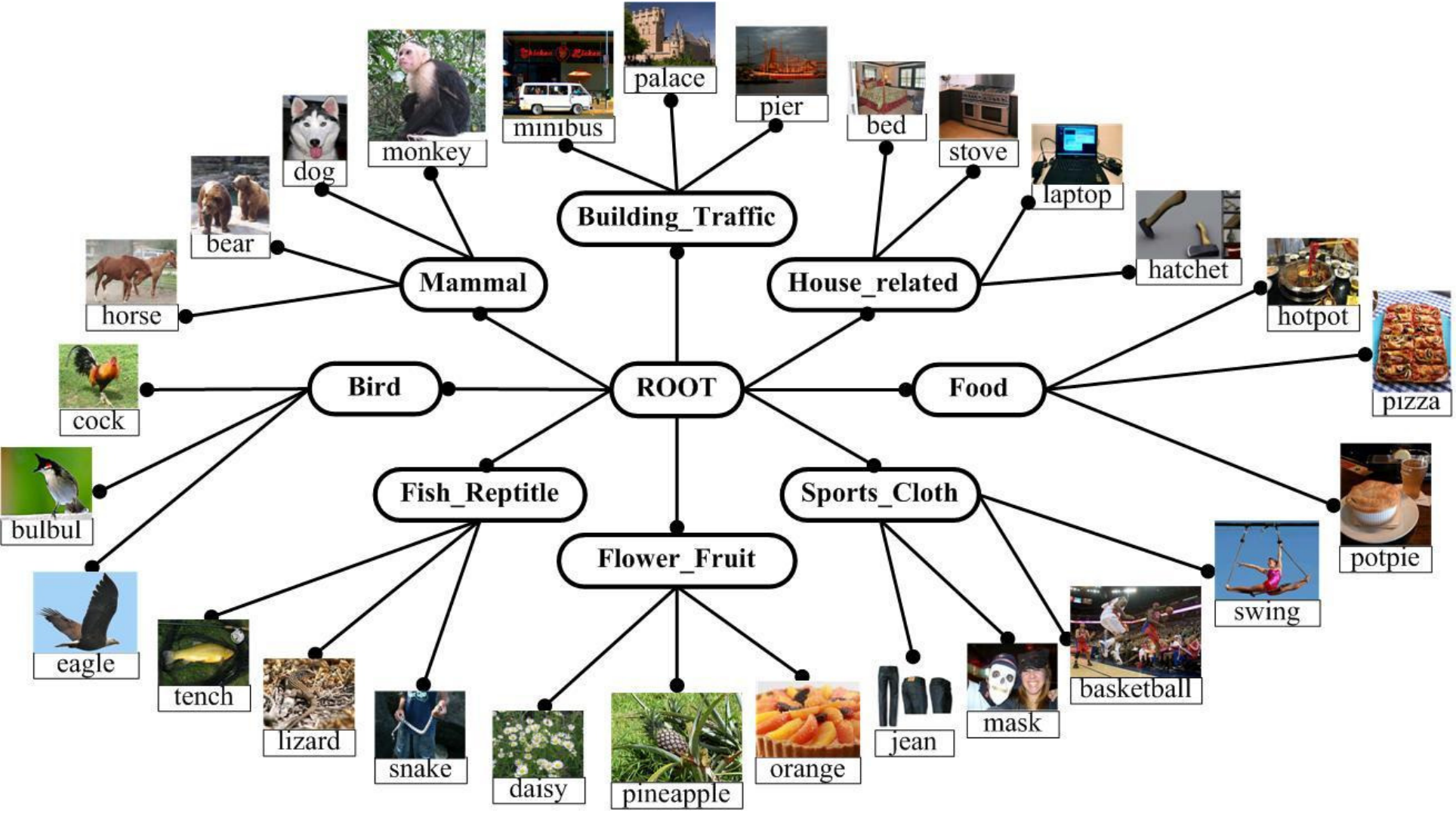}
\end{center}
\vspace*{-0.488cm}
   \caption{\bf Part of our two-layer ontology for indexing $1,000$ atomic object classes in ImageNet1K image set.}
\label{fig:long}
\label{fig:onecol}
\vspace*{-0.318cm}
\end{figure}

Given these $7,756$ atomic object classes, we are interested in learning a two-layer (category layer and object class layer) ontology that comprises: (a) one root node that contains all these $7,756$ atomic object classes; (b) a set of non-leaf nodes (category nodes) and each of them contains one particular subset of $7,756$ atomic object classes; (c) $7,756$ leaf nodes and each of them contains one particular atomic object class; and (d) a set of edges to illustrate hierarchical inter-node relationships. Our algorithm for two-layer ontology construction contains two key components: (1) estimating the inter-class semantic relationships and generating an inter-class semantic relationship matrices $\mathbf{\Psi}$; (2) partitioning $7,756$ atomic object classes into a set of categories according to their inter-class semantic relationships.  

For two given atomic object classes $c_i$ and $c_j$, their inter-class semantic relationship $\psi_{i,j}$ is defined as:  
$$\psi_{i,j} = \psi(c_i, c_j) = - \log\frac{D(c_i, c_j)}{2H} \eqno(1)$$
where $D(c_i, c_j)$ is the number of nodes to be traveled from the concept node for $c_i$ to the concept node for  $c_j$ over the WordNet [71-74], $H$ is the maximum number of the nodes to be traveled from the root node to the deepest leaf node. Finally, the inter-class semantic relationship matrix $\mathbf{\Psi}$ for $7,756$ atomic object classes is obtained and its component is determined as $\psi_{i,j}$. 

Spectral clustering is then performed on the semantic relationship matrix $\mathbf{\Psi}$ by partitioning $\mathbf{\Psi}$ into a set of small blocks, and each small block corresponds to one particular category node which is associated with a set of semantically-related atomic object classes. Such semantically-related atomic object classes, which are associated with the same category node, are directly assigned into a set of sibling leaf nodes and each leaf node contains one particular atomic object class. Fig. 1 is used to illustrate our experimental result on two-layer ontology construction for ImageNet10K image set. Our algorithm can also be applied to ImageNet1K image set, and the experimental result on two-layer ontology construction is shown in Fig. 2. 

For the semantically-related atomic object classes on the sibling leaf nodes, the tasks for learning their deep CNNs and softmax are strongly inter-related, thus our two-layer ontology can provide a good environment to determine the inter-related learning tasks automatically for supporting multi-task learning. By grouping the semantically-related atomic object classes into the same category node and assigning them onto the sibling leaf nodes, our two-layer ontology can provide a good environment for task group generation by assigning the semantically-related atomic object classes on the sibling leaf nodes into the same task group (i.e., the same base deep CNNs).

\section{Deep Mixture of Diverse Experts} 
In this paper, a deep mixture of diverse experts algorithm is developed to recognize tens of thousands of atomic object classes by seamlessly combining a set of base deep CNNs with diverse outputs (task spaces): (a) The network structure (number of layers and number of units in each layer) of the well-designed AlexNet [11-13] is directly used to configure the base deep CNNs with $M+1$ ($M \leq 1,000$) outputs (i.e., for $M$ atomic object classes and one special class of ``not-in-group" in the current task group); (b) The training images for $M$ atomic object classes in the current task group and one special class of ``not-in-group" are used to fine-tune and optimize the node weights for the corresponding base deep CNNs; (c) All these base deep CNNs are trained to recognize different subsets of tens of thousands of atomic object classes rather than the same set of atomic object classes. 

First, our two-layer ontology is used to achieve more effective solution for task group generation, e.g., the semantically-related atomic object classes on the sibling leaf nodes are assigned into the same task group and each task group contains $M$ ($M \leq 1,000$) atomic object classes and one specific class of ``not-in-group". In this tree-guided task assignment process, inter-group overlapping is allowed for supporting message passing among the task groups (i.e., base deep CNNs), so that the diverse predictions from all these base deep CNNs may become more comparable. Second, one particular base deep CNNs with $M+1$ outputs (for recognizing $M$ atomic object classes and identifying one specific class of ``not-in-group") is learned for each task group and deep multi-task learning is performed to leverage the inter-class visual similarities to learn more discriminative base deep CNNs and multi-task softmax for enhancing the separability of the atomic object classes in the same task group. Finally, all these base deep CNNs with diverse outputs (task spaces) are seamlessly integrated to form a deep mixture of diverse experts for recognizing tens of thousands of atomic object classes effectively.

\subsection{Tree-Guided Task Assignment}
It is worth noting that our two-layer ontology can effectively assign the semantically-related object classes into the sibling leaf nodes. As illustrated in Fig. 3, our two-layer ontology is used to guide the left-to-right task assignment process for task group generation by assigning the semantically-related object classes on the sibling leaf nodes into the same task group, so that their base deep CNNs and multi-task softmax can be learned jointly in an end-to-end fashion to enhance their separability. 
\begin{figure}[t]
\vspace*{-0.188cm}
\begin{center}
\includegraphics[width=0.48\textwidth]{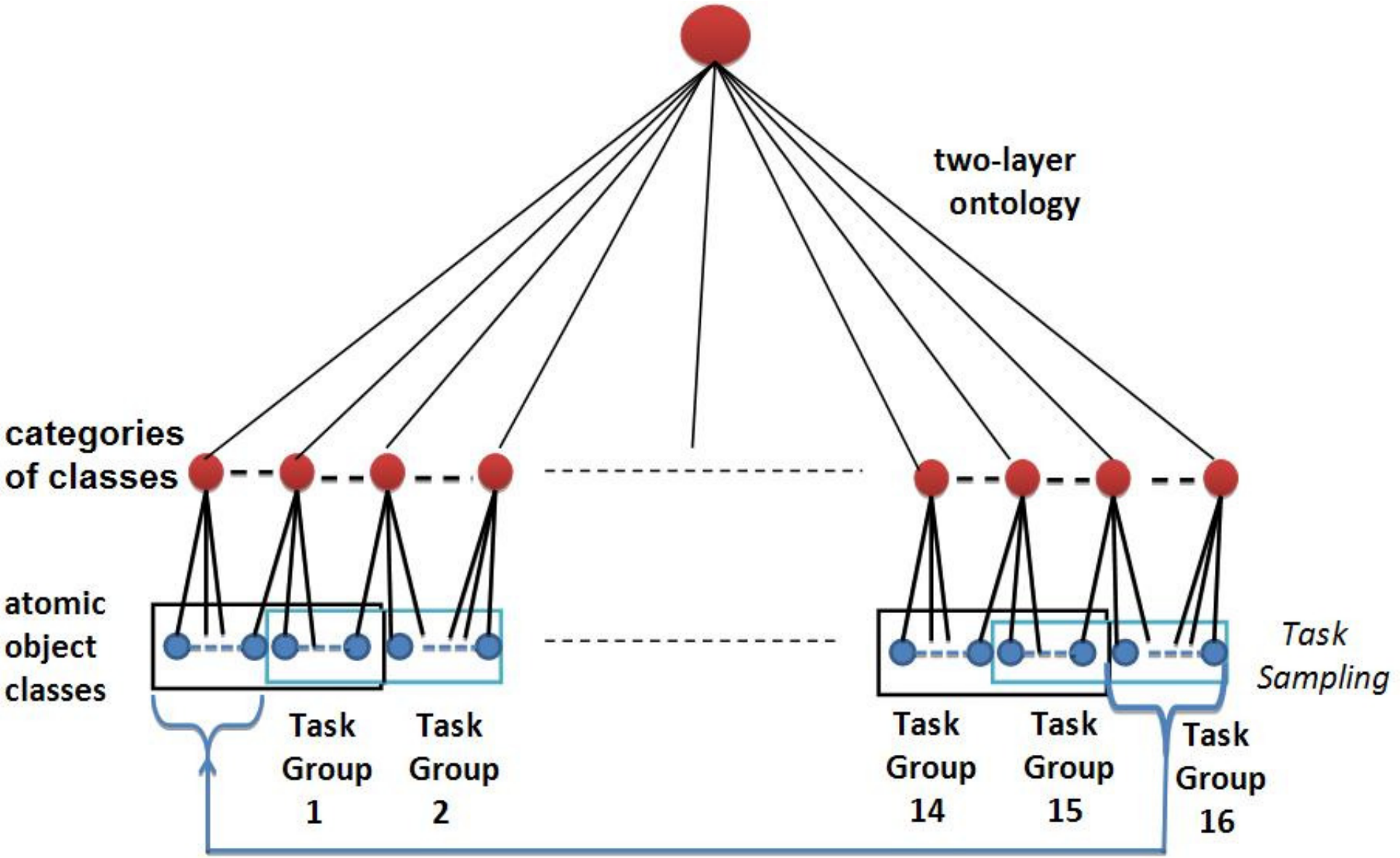}
\end{center}
\vspace*{-0.488cm}
   \caption{\bf The flowchart to illustrate the tree-guided task assignment process for task group generation when 50\% inter-group overlapping is used. }
\label{fig:long}
\label{fig:onecol}
\vspace*{-0.188cm}
\end{figure}

Our goal is to combine a set of base deep CNNs with diverse outputs to recognize $7,756$ atomic object classes in ImageNet10K, thus each task group contains $M$ ($M \leq 1,000$) atomic object classes and one specific class of ``not-in-group" is added. Such special class of ``not-in-group" is used to: (a) establish inter-group correlations and support inter-group message passing to make their predictions to be comparable, e.g., if the corresponding atomic object class for a given image or object proposal does not belong to the current task group, the special class of ``not-in-group" in the current task group may obtain the maximum prediction score, thus such special class of ``not-in-group" can be used to pass necessary messages among the task groups and the predictions for the atomic object classes in the current task group are less reliable or have less contributions on the 7,756-D high-level features for image semantics representation; (b) avoid unreliable predictions by assigning the maximum prediction score for the special class of ``not-in-group" rather than distributing the prediction scores uniformly over all the irrelevant atomic object classes in the current task group. 

By integrating our two-layer ontology to guide the left-to-right task assignment process for task group generation, the semantically-related atomic object classes on the sibling leaf nodes can be assigned into the same task group, so that our deep multi-task learning algorithm can leverage their inter-class visual similarities to learn their base deep CNNs and multi-task softmax jointly in an end-to-end fashion to enhance their separability.  If 50\% inter-group overlapping is allowed, each atomic object class can be assigned into two task groups, thus we can finally generate $16$ task groups for recognizing $7,756$ atomic object classes in ImageNet10K image set. By allowing inter-group task overlapping and adding one specific class of ``not-in-group" in each base deep CNNs, our deep mixture of diverse experts algorithm can effectively support message passing among these base deep CNNs and make their predictions to be comparable (even they are not completely trained jointly).

\subsection{Deep Multi-Task Learning for Each Task Group}
For a given task group with $M$ atomic object classes ($M \leq 1,000$), one particular base deep CNNs is learned and the network structure of the well-designed AlexNet [11-13] is directly used to configure such base deep CNNs with $M+1$ outputs (for recognizing $M$ atomic object classes and identifying one specific class of ``not-in-group"). For $M$ atomic object classes in the same task group ($M \leq 1,000$), the model parameters for their multi-task softmax are trained simultaneously by optimizing a joint objective function:
$$min \left\{\mu \sum_{l=1}^R\sum_{j=1}^{M} \xi^l_j + \delta_1 Tr\left(WW^T\right) + \frac{\delta_2}{2}Tr\left(WLW^T\right)\right\}  \eqno(2)$$ 
where $R$ is the number of training images for each atomic object class, $Tr(\cdot)$ is used to represent the trace of matrix, $\xi^l_j$ indicates the training error rate, $\delta_1$ and $\delta_2$ are the regularization parameters, $\mu$ is the penalty term, $W = (W_1, \cdots, W_j, \cdots, W_M)$ is the set of the multi-task model parameters for $M$ atomic object classes in the same task group, $L$ is the Laplacian matrix of the relevant inter-class visual similarity matrix $\mathbf{S}$.

The $M \times M$ visual similarity matrix $\mathbf{S}$ is used to characterize the inter-class visual similarities for $M$ atomic object classes in the same task group and its component $S_{ij}$ is defined as: 
$$S_{ij} = S(c_i, c_j) =\frac{1}{R^2}\sum_{l=1}^R\sum_{m=1}^R \kappa(x^i_l, x^j_m) \eqno(3)$$
where $\kappa(\cdot, \cdot)$ is the kernel function for visual similarity characterization, $x^i_k$ and $x^j_m$ are the deep features for the $l$th image from the $i$th atomic object class $c_i$ and the $m$th image from the $j$th atomic object class $c_j$. The deep features provided by AlexNet [11-13] are used to initialize such inter-class visual similarity matrix $\mathbf{S}$ and the newly-learned deep features (obtained by our base deep CNNs) are further used to update such inter-class similarity matrix $\mathbf{S}$ iteratively. 

The inter-class visual similarities are used to approximate the inter-task correlations, the manifold regularization term $Tr\left(WLW^T\right)$ is used to enforce that: if two atomic object classes $c_i$ and  $c_j$ have larger inter-class visual similarity, their multi-task model parameters $W_i$ and $W_j$ may share some common components significantly. Our deep multi-task learning algorithm takes the following operations to enhance the separability of the atomic object classes in the same task group: (1) explicitly separating the group-wise common prediction component $W_0$ from the class-specific discrimination component $V_j$: $W_j = W_0 + V_j$, $j \in \{1, \cdots, M\}$; (2) leveraging their inter-class visual similarities to approximate their inter-task correlations and optimize their multi-task model parameters $W = W_0 + V_j$, $j \in \{1, \cdots, M\}$ jointly; and (3) using the class-specific discrimination components $V_j$, $j \in \{1, \cdots, M\}$ to separate the atomic object classes in the same task group rather than paying attention on their group-wise common prediction component $W_0$. 

By embedding the inter-class visual similarities (i.e., inter-task correlations) into the manifold structure regularization term, our deep multi-task learning algorithm can learn more discriminative base deep CNNs and multi-task softmax for the atomic object classes in the same task group. By comparing and contrasting such atomic object classes in the same task group simultaneously, our deep multi-task learning algorithm can optimize their base deep CNNs and multi-task softmax jointly to enhance their separability even such semantically-related atomic object classes are usually hard to be distinguished. Because the atomic object classes in the same task group may have strong inter-class correlations and share similar learning complexities, the gradients of their joint objective function could be more uniform and the back-propagation operations can stick on reaching the global optimum effectively. Thus our deep multi-task learning algorithm can obtain more discriminative base deep CNNs and multi-task softmax for enhancing the separability of the atomic object classes in the same task group.

We simultaneously optimize the multi-task model parameters for the atomic object classes in the same task group according to their joint objective function as defined in Eq. (2), and the errors are further back-propagated to refine the weights for the base deep CNNs. Given a training image, the predictions of the atomic object classes in the same task group are calculated. We formulate the training error rate $\xi^l_j$ in the form of softmax regression:
$$\xi^l_j =  -\mathds{I}\{y^l_j\}log\left\{\frac{exp(W_j^Tx^l_j+b)}{\sum_{i=1}^{M+1}exp(W_i^Tx^l_i+b)}\right\}\eqno(4)$$
where $\mathds{I}\{y^l_j\}$ is the indicator function such that $\mathds{I}\{y^l_j\} = 1$ if $y^l_j = 1$ (i.e., $(x^l_j,y^l_j)$ is the positive training image for the atomic object class $c_j$), otherwise $\mathds{I}\{y^l_j\} = 0$. The joint objective function in Eq.(2) is then reformulated as:
{\small $$ \pounds(W,X,Y) = -\mu \sum_{l=1}^R\sum_{j=1}^{M} \mathds{I}\{y^l_j\}log\left\{\frac{exp(W_j^Tx^l_j+b)}{\sum_{i=1}^{M+1}exp(W_i^Tx^l_i+b)}\right\} $$}
{\small $$ + \delta_1 Tr\left(WW^T\right) + \frac{\delta_2}{2}Tr\left(WLW^T\right) \eqno(5)$$

The corresponding gradients for the joint objective function $\pounds(W,X,Y)$ in Eq. (5) are calculated as $\frac{\partial \pounds(W,X,Y)}{\partial W}$: 
$$ \frac{\partial \pounds(W,X,Y)}{\partial W_j} = -\mu  \sum_{l=1}^{\mathds{I}\{y^l_j\}} x_l \left\{ 1- \frac{exp(W_j^Tx^l+b)}{\sum_{i=1}^{M+1}exp(W_i^Tx^l+b)}\right\}$$ $$+\delta_1 W_j + \delta_2W_jL \eqno(6)$$ 
Such gradients are back-propagated [18] through the base deep CNNs to fine-tune the node weights and the set of multi-task model parameters $W$.

Because the inter-class visual similarities are explicitly considered in the manifold regularization term $Tr\left(WLW^T\right)$ and in the joint objective function $\pounds(W,X,Y)$, back-propagating the gradients of the joint objective function $\frac{\partial \pounds(W,X,Y)}{\partial W}$ to fine-tune the node weights for the base deep CNNs can allow us to learn more discriminative base deep CNNs and multi-task softmax for enhancing the separability of the atomic object classes in the same task group. The newly-learned deep features are further used to update the inter-class similarity matrix $\mathbf{S}$ iteratively. 
\begin{figure}[t]
\vspace*{-0.188cm}
\begin{center}
\includegraphics[width=0.48\textwidth]{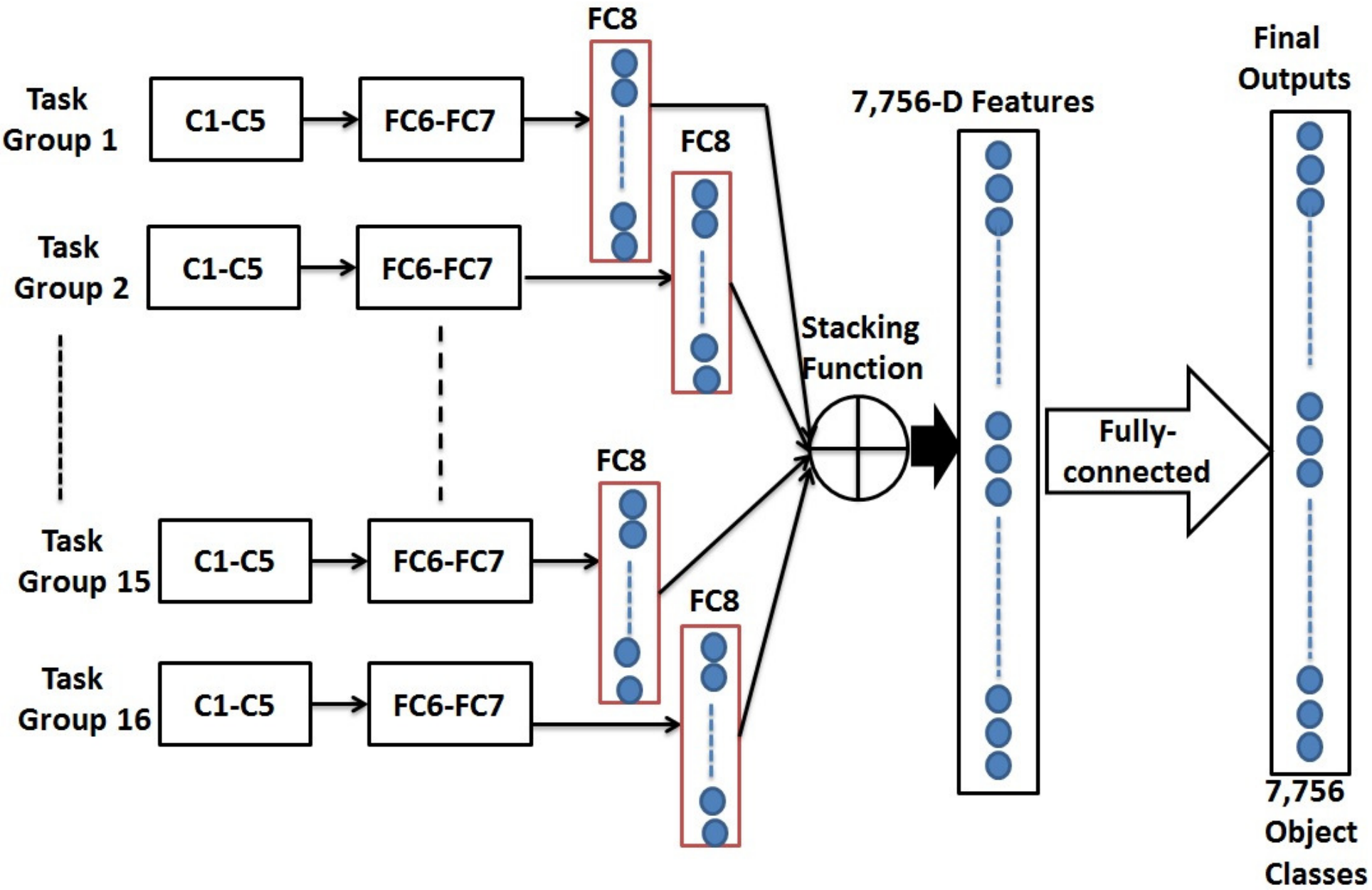}
\end{center}
\vspace*{-0.488cm}
   \caption{\bf The flowchart of our deep mixture of diverse experts algorithm when 50\% inter-group overlapping is used for task group generation. }
\label{fig:long}
\label{fig:onecol}
\vspace*{-0.188cm}
\end{figure}

\subsection{Stacking Function for Fusing Diverse Predictions} 
As illustrated in Fig. 4, our deep mixture of diverse experts algorithm uses a stacking function to combine the diverse outputs from a set of base deep CNNs to generate 7,756-D high-level features for image semantics representation, e.g., like object bank [36-38], we use the prediction scores for the appearances of 7,756 atomic object classes (from all these base deep CNNs) to generate the high-level features for image semantics representation. Such 7,756-D high-level features are fully connected with a softmax with 7,756 outputs (i.e., for recognizing 7,756 atomic object classes). In this work, we have leveraged three factors to design the stacking function to extract such 7,756-D high-level features for image semantics representation: 

(a) {\bf\em Prediction Scores:} For a given image or object proposal, even its corresponding atomic object class does not appear in all these base deep CNNs or its labeled object class may just appear in few base deep CNNs, all other irrelevant base deep CNNs may still provide their individual predictions. However, only the predictions from few relevant base deep CNNs can properly be integrated to indicate the appearance of the corresponding atomic object class for the given image or object proposal, thus the predictions from other irrelevant base deep CNNs should have low contributions on the 7,756-D high-level features for image semantics representation. 

(b) {\bf\em Inter-Group Conflict:} If a given image or object proposal receives conflict predictions from different base deep CNNs, it is reasonable for our stacking function to assign lower prediction scores for the appearances of the corresponding atomic object classes in its 7,756-D high-level features. The prediction scores for the specific class of ``not-in-group" can be used to identify such inter-group conflict effectively, e.g., if the given image or object proposal receives the maximum prediction score for the special class of ``not-in-group" in the current base deep CNNs, all the atomic object classes in the current base deep CNNs should have lower appearance scores in its 7,756-D high-level features for image semantics representation.  

(c) {\bf\em Inter-Group Overlapping Percentage:} When more inter-group overlapping percentages are allowed, each atomic object class can be assigned into more base deep CNNs and receive more sufficient comparison and contrasting with others from different aspects, e.g., the atomic object class for the given image or object proposal may appear in more base deep CNNs, thus such inter-group overlapping percentage $\lambda$ may have significant impacts on its 7,756-D high-level features for image semantics representation. 

Based on these understandings, for a given image or object proposal, the $i$th component $\Upsilon(i)$ on its 7,756-D high-level features is used to indicate the appearance probability of the $i$th atomic object class $c_i$  and it is defined as: 
$$\Upsilon(i) =  \sum_{j=1}^{\vartheta} \Lambda_j (c_i) PS(i,j) \frac{(1- \phi_j)}{\phi_j}  \eqno(7)$$ 
where $\vartheta$ is the total number of base deep CNNs being combined, $\phi_j$ is the prediction score for the special class of ``not-in-group" in the $j$th base deep CNNs,  $\Lambda_j (c_i)$ is an indication function to characterize the appearance of the atomic object class $c_i$ in the $j$th base deep CNNs, $PS(i,j)$ is the prediction score for the given image or object proposal to be assigned into the $i$th atomic object class $c_i$ by the $j$th base deep CNNs. 

The indication function $\Lambda_j (c_i)$ is defined as: 
$$
\Lambda_j(c_i) = 
  \begin{cases} 
   1,	 & \text{if } c_i \text{ is in the jth task group }  \\
   \\
   \lambda,      & \text{otherwise }
  \end{cases}
\eqno(8)$$
where $0 \leq \lambda \leq 1$ is the inter-group overlapping percentage. When $\lambda = 1$, all these base deep CNNs share the same task spaces or outputs (i.e., they are trained to recognize the same set of $1,000$ atomic object classes), thus our deep mixture of diverse experts algorithm is degraded to the traditional approaches for combining deep CNNs [40-52]. When $\lambda = 0$, all these base deep CNNs have totally-different task spaces without inter-group overlapping, thus only $8$ base deep CNNs are trained for recognizing $7,756$ atomic object classes. It is worth noting that the total number $\vartheta$ of base deep CNNs being fused is implicitly depended on the inter-group overlapping percentage $\lambda$ for task group generation. 

For a given image or object proposal, $PS(i,j)$ is used to indicate its prediction score to be assigned into the $i$th atomic object class $c_i$ by the $j$th base deep CNNs: 
$$
PS(i,j) = 
  \begin{cases} 
   p_j(i),	 & \text{if } c_i \text{ in the jth task group }  \\
   \\
   0,      & \text{otherwise }
  \end{cases}
\eqno(9)$$
$$p_j(i) = \frac{exp(W_i^T x + b)}{\sum_{l=1}^{M+1}exp(W_l^T x + b)}$$
where $p_j(i)$ is the $i$th output (for the $i$th atomic object class $c_i$) in the $j$th base deep CNNs, $M \leq 1,000$ is the total number of atomic object classes in the $j$th base deep CNNs, $0 \leq p_j(i) \leq 1$  is the prediction score for the $i$th atomic object class $c_i$ that is provided by the $j$th base deep CNNs when the $i$th atomic object class $c_i$ appears in the $j$th task group (base deep CNNs), otherwise,  $ p_j(i) = 0$ when the $i$th atomic object class $c_i$ does not belong to the $j$th task group (base deep CNNs). 

Like object bank [36-38], we integrate the prediction scores $\Upsilon$ for all these 7,756 atomic object classes to generate the 7,756-D high-level features for image semantics representation, such 7,756-D high-level features are fully connected with a 7,756-way softmax. The objective function for the 7,756-way softmax is defined as:
$$
\pounds(y, x) = \sum_{m=1}^R\sum_{h=1}^{\Omega}\mathds{I}\{y^m_h\}log\left\{\frac{exp(W_h^Tx^m_h)}{\sum_{g=1}^{\Omega}exp(W_g^Tx^m_g)}\right\}
\eqno(10)$$
where $R$ is the number of training images, $\Omega = 7,756$ is the total number of atomic object classes being recognized, $\mathds{I}\{y\}$ is the identification function. The gradients of the joint objective function $\frac{\partial \pounds(y, x)}{\partial W}$ are calculated as: 
$$ \frac{\partial \pounds(y, x)}{\partial W_h} = -\mu  \sum_{m=1}^{\mathds{I}\{y^m_h\}} x^m \left \{ 1- \frac{exp(W_h^Tx^m+b)}{\sum_{i=1}^{\Omega}exp(W_i^Tx^m+b)}\right \} \eqno(11)$$
Such gradients are further back-propagated to fine-tune: (1) The model parameters for the 7,756-way softmax at the stacking level; (2) The model parameters for the $(M+1)$-way ($M \leq 1,000$) multi-task softmax at the base level (at the task group level); (3) The node weights for the relevant base deep CNNs.

\section{Experimental Results} 
We have evaluated our deep mixture of diverse experts algorithm over ImageNet10K image set with 10,184 image categories [24], and $7,756$ atomic object classes are identified. We have compared our deep mixture of diverse experts algorithm with the state-of-the-art baseline methods and our comparison experiments focus on evaluating the following factors: (a) whether the number of base deep CNNs being combined and the inter-group overlapping percentage being used have significant impacts on improving the performance of our deep mixture of diverse experts algorithm; (b) whether our deep mixture of diverse experts approach can achieve higher accuracy rates on large-scale visual recognition; and (c) whether our deep multi-task learning algorithm can effectively leverage the inter-class visual similarities to learn more discriminative base deep CNNs and multi-task softmax for enhancing the separability of the atomic object classes in the same task group. 
\begin{figure}[t]
\begin{center}
\includegraphics[width=0.48\textwidth]{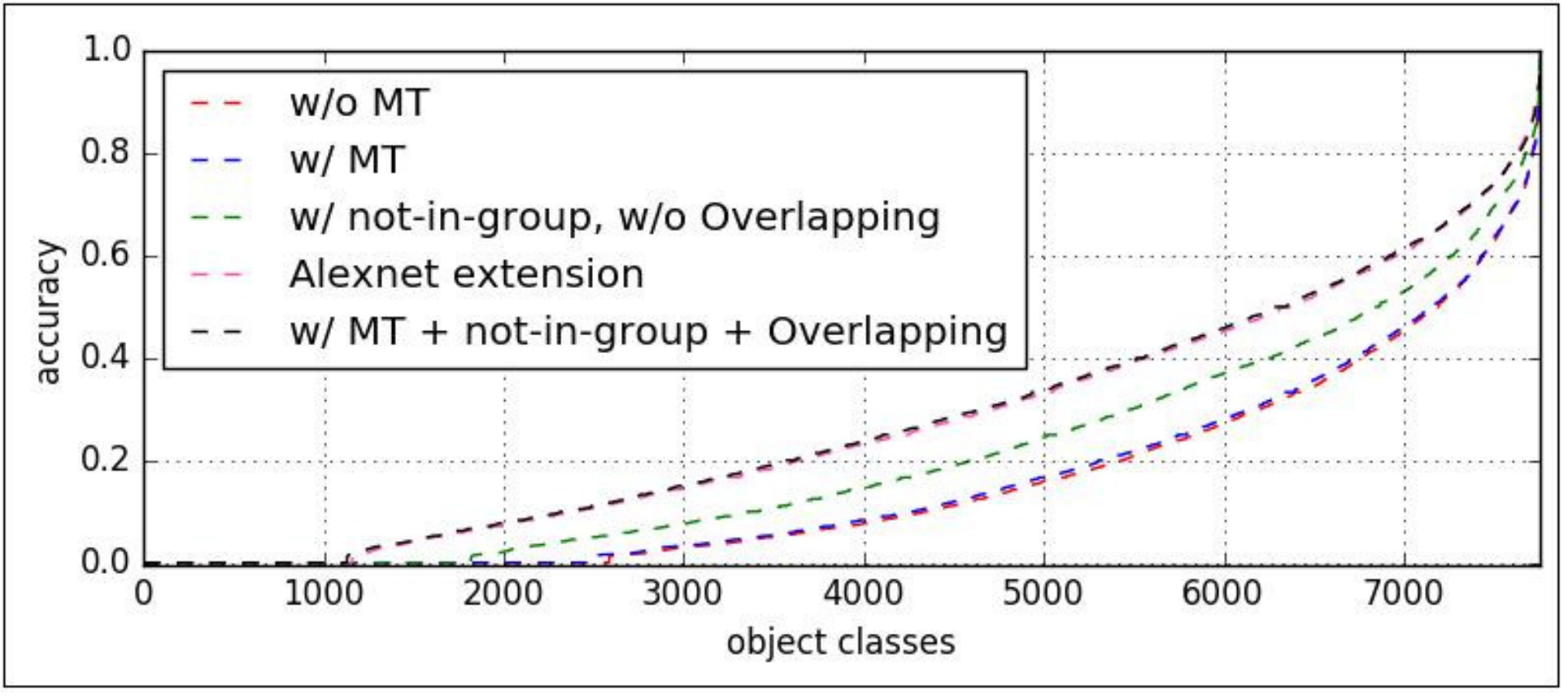}
\end{center}
\vspace*{-0.388cm}
   \caption{\bf The comparison on the accuracy rates for: (a) our deep mixture of diverse experts algorithm when deep multi-task learning, inter-group overlapping and special class of ``not-in-group" are used; (b) AlexNet Extension; (c) our deep mixture algorithm without multi-task learning; (d) our deep mixture of diverse experts algorithm without using inter-group overlapping but having special class of ``not-in-group"; (e) our deep mixture of diverse experts algorithm with only multi-task learning.  }
\label{fig:long}
\label{fig:onecol}
\vspace*{-0.188cm}
\end{figure}
\begin{figure}[t]
\begin{center}
\includegraphics[width=0.48\textwidth]{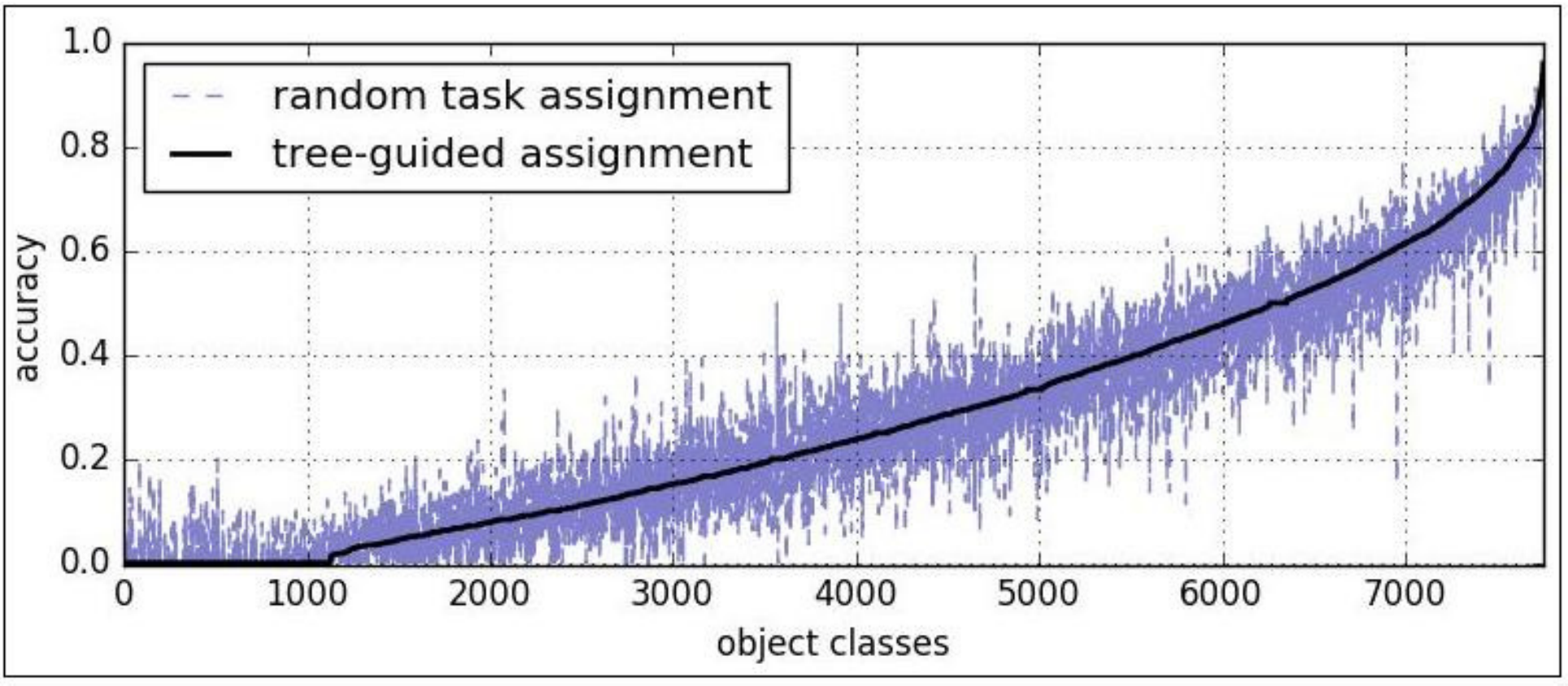}
\end{center}
\vspace*{-0.388cm}
   \caption{\bf The comparison on the accuracy rates for our deep mixture of diverse experts algorithm: (a) tree-guided task assignment is used; (b) random task assignment is used.  }
\label{fig:long}
\label{fig:onecol}
\vspace*{-0.188cm}
\end{figure}

{\bf (a) Effectiveness of our deep mixture of diverse experts algorithm:} To evaluate the effectiveness of our deep mixture of diverse experts algorithm, we have compared: (1) {\em AlexNet Extension}: we simply enlarge the FC8 layer of AlexNet [11-13] from a 1,000-way softmax to a 7,756-way softmax; (2) {\em Random task assignment}: all these 7,756 atomic object classes are randomly assigned into multiple task groups without considering their inter-class correlations and each task group contains $M$ ($M \leq 1,000$) randomly-selected atomic object classes; (3) {\em Under various conditions}: whether various operations, such as performing deep multi-task learning, allowing inter-group overlapping and adding special class of ``not-in-group", have significant impacts on the performances of our deep mixture of diverse experts algorithm. As shown in Fig. 5 and Fig. 6, one can easily observe that our deep mixture of diverse experts algorithm can achieve better performance on large-scale visual recognition. The comparisons on their average accuracy rates are shown in Table I. 
\begin{table}
\caption{\bf The comparisons on the average accuracy rates.}
\begin{center}
\begin{tabular}{|l|c|c|c|}
\hline
\multirow{2}{*}{approaches} & 
\multicolumn{3}{|c|}{accuracy rate (top k)}
\\
\cline{2-4}& 1  & 5 & 10 
\\
\hline 
  our deep mixture algorithm						& \textbf{38.65\%}	 &\textbf{55.41\%} &\textbf{64.32\%} 	\\
  \hline
  AlexNet Extension					& 31.70\% 		& 46.23\% 	&52.18\%\\ 
\hline 
 random  assignment              & 34.53\%	 & 47.39\% & 53.25\% 	\\
\hline 
 visual tree                 & 37.55\%	& 53.29\% & 62.02\% 	\\
\hline 
 Stack 2                & 37.63\%	& 54.37\% & 63.29\% 	\\ 
 \hline 

\end{tabular}
\end{center}\label{tab:10k}
\end{table}

As shown in Fig. 5, for most of 7,756 atomic object classes in ImageNet10K, our deep mixture of diverse experts algorithm can achieve higher accuracy rates. The reasons are: (1) Combining a set of base deep CNNs with diverse outputs can allow each atomic object class to receive more sufficient comparison and contrasting with others from different aspects and obtain multiple predictions from the relevant base deep CNNs, which may significantly increase its chances to be recognized correctly. (2) Our tree-guided task assignment algorithm can assign the semantically-related atomic object classes with similar learning complexities into the same task group, so that our deep multi-task learning algorithm can leverage their inter-class visual similarities to learn more discriminative base deep CNNs and multi-task softmax jointly in an end-to-end fashion for enhancing their separability significantly. (3) By adding one special class of ``not-in-group" in each base deep CNNs and allowing inter-group overlapping, our deep mixture of diverse experts algorithm can effectively support message passing among the base deep CNNs and make their predictions being more comparable. By passing necessary messages among the base deep CNNs, we can integrate their diverse predictions to generate 7,756-D high-level features for image semantics representation even these base deep CNNs are not completely trained jointly. As shown in Fig. 6, one can observe that our deep mixture of diverse experts algorithm with tree-guided task assignment can have better performance than that when random task assignment is used. 

The comparisons on the average accuracy rates are given in Table I, one can easily observe that: (1) our deep mixture of diverse experts algorithm can achieve better performance than the AlexNet Extension approach; and (2) our deep mixture of diverse experts algorithm can achieve better performance when tree-guided task assignment is used. As shown in Fig. 7, Fig. 8 and Fig. 9, we have also demonstrated the prediction scores, one can observe that our deep mixture of diverse experts algorithm can provide higher prediction scores than the AlexNet Extension approach does. 
\begin{figure}[t]
\vspace*{-0.188cm}
\begin{center}
\includegraphics[width=0.48\textwidth]{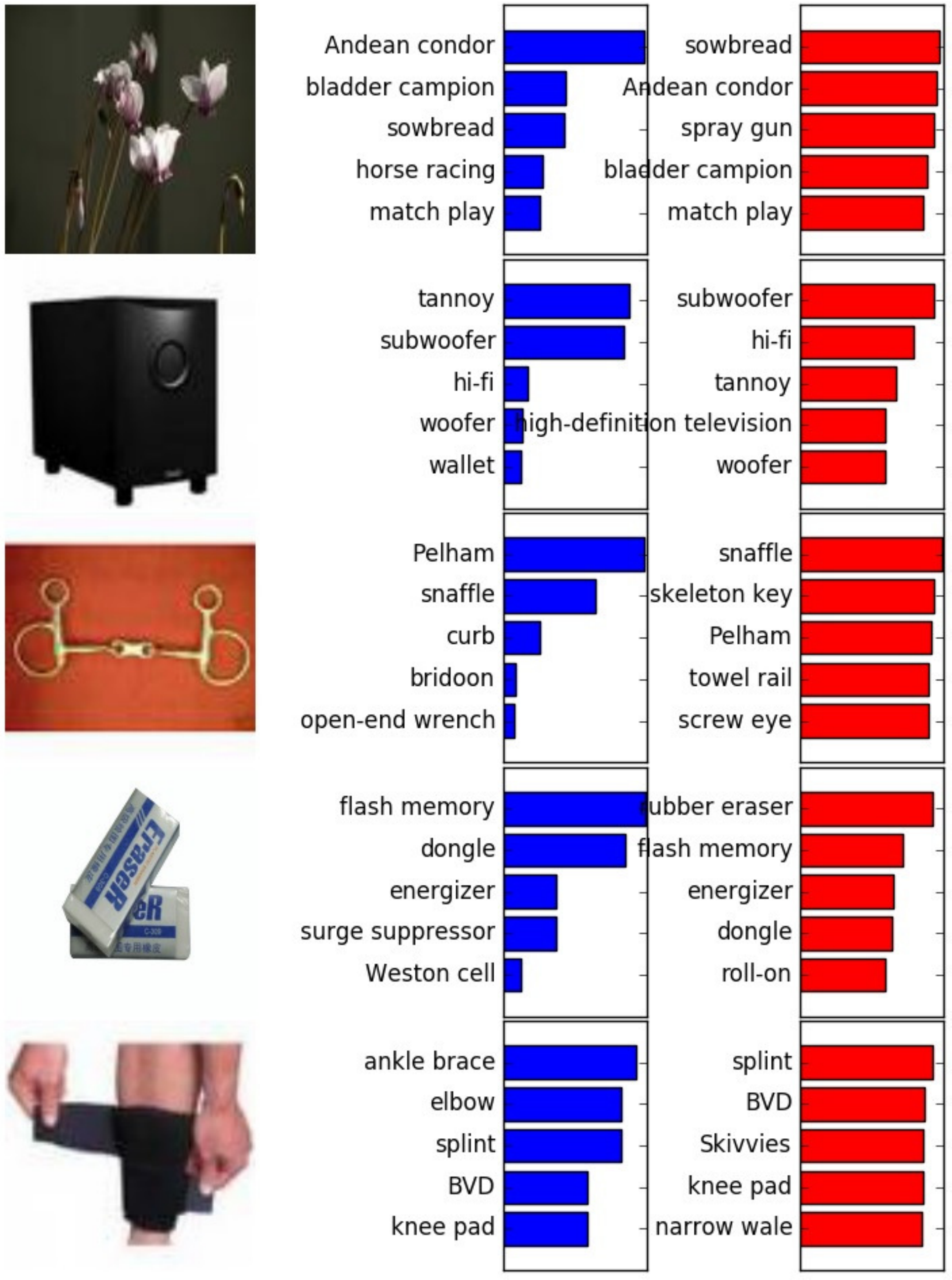}
\end{center}
\vspace*{-0.488cm}
   \caption{\bf The comparisons on the prediction scores: (a) test images; (b) predicted object classes and their scores from our deep mixture of diverse experts algorithm; (c) predicted object classes and their scores from the AlexNet Extension approach. }
\label{fig:long}
\label{fig:onecol}
\vspace*{-0.188cm}
\end{figure}
\begin{figure}[t]
\vspace*{-0.188cm}
\begin{center}
\includegraphics[width=0.48\textwidth]{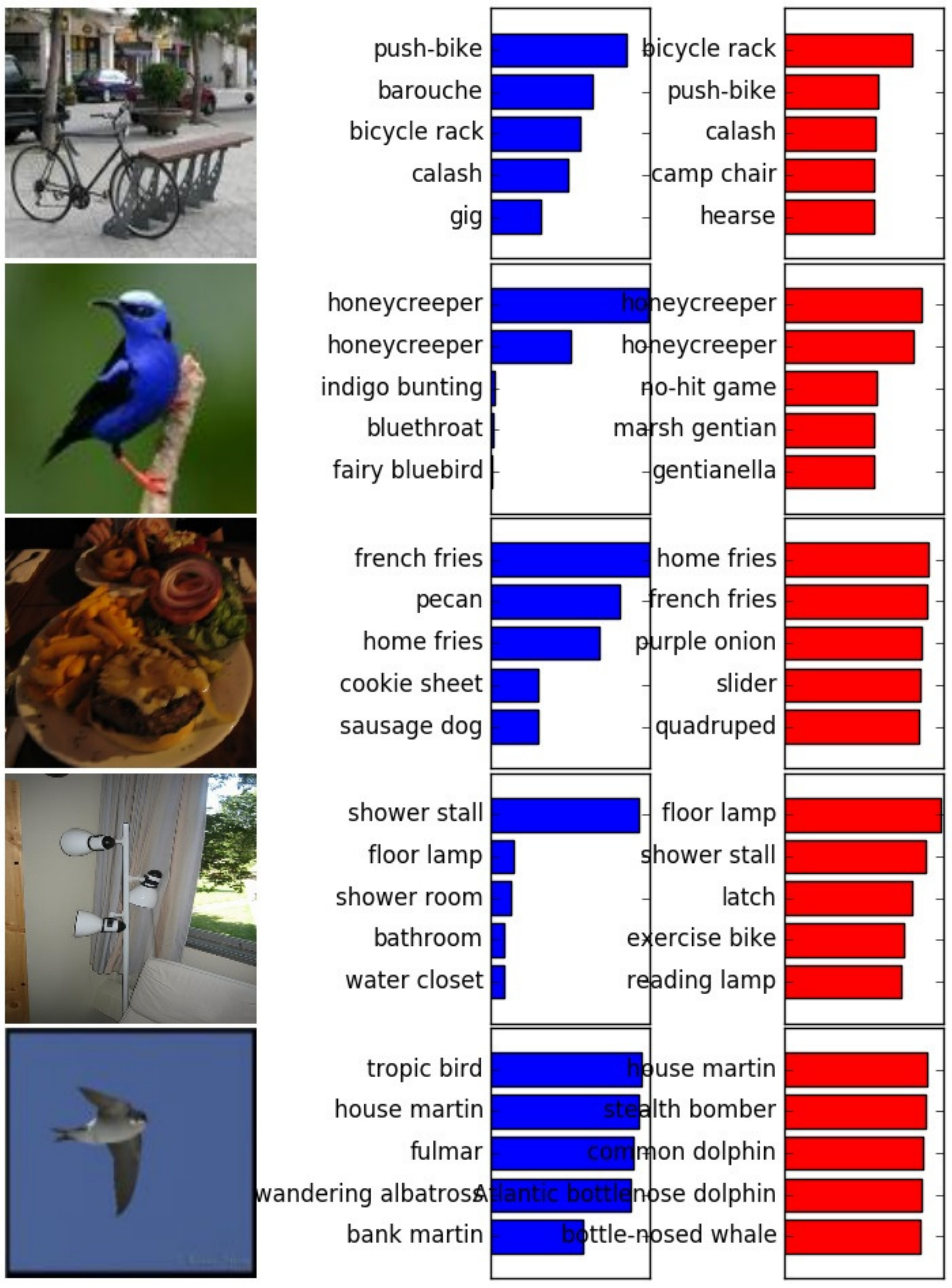}
\end{center}
\vspace*{-0.488cm}
   \caption{\bf The comparisons on the prediction scores: (a) test images; (b) predicted object classes and their scores from our deep mixture of diverse experts algorithm; (c) predicted object classes and their scores from the AlexNet Extension approach. }
\label{fig:long}
\label{fig:onecol}
\vspace*{-0.188cm}
\end{figure}
\begin{figure}[t]
\vspace*{-0.288cm}
\begin{center}
\includegraphics[width=0.48\textwidth]{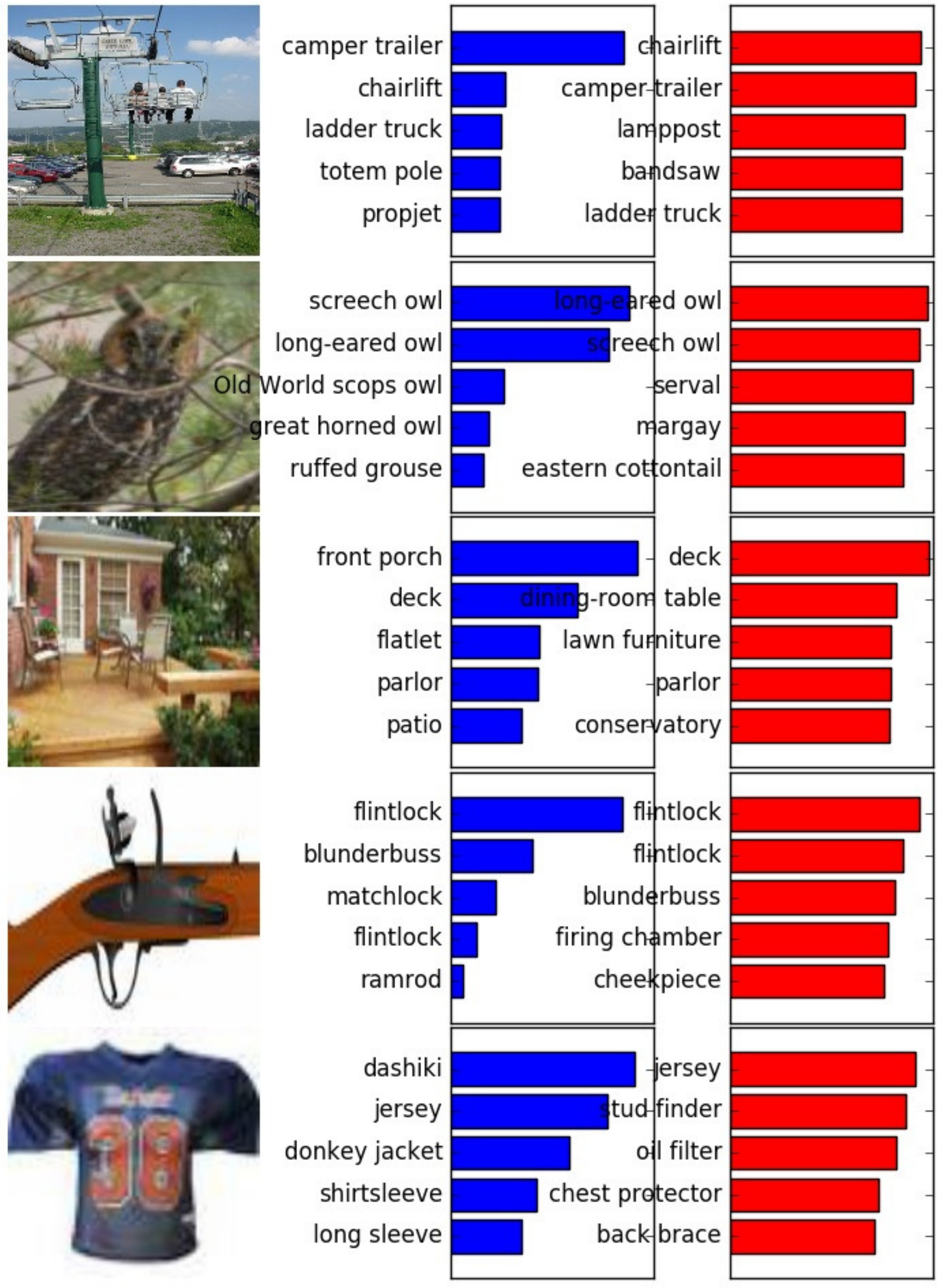}
\end{center}
\vspace*{-0.488cm}
   \caption{\bf The comparisons on the prediction scores: (a) test images; (b) predicted object classes and their scores from our deep mixture of diverse experts algorithm; (c) predicted object classes and their scores from the AlexNet Extension approach. }
\label{fig:long}
\label{fig:onecol}
\vspace*{-0.188cm}
\end{figure}

{\bf (b) Effectiveness of deep multi-task learning:} 
To evaluate the effectiveness of our deep multi-task learning algorithm at the group level, we have compared it with the baseline approach AlexNet [11-13] (i.e., without considering the inter-task correlations). For the atomic object classes in the same task group, our deep multi-task learning algorithm can explicitly leverage their inter-class visual similarities (through the Laplacian matrix $L$) to learn more discriminative base deep CNNs and multi-task softmax jointly in an end-to-end fashion. As shown in Fig. 10, Fig. 11, Fig. 12 and Table II, our deep multi-task learning algorithm can obtain higher accuracy rates than the baseline method AlexNet [11-13], where the atomic object classes in the same task group are sorted according to their accuracy rates that are obtained by our deep multi-task learning algorithm. Because the atomic object classes in the same task group share similar learning complexities, the gradients of their joint objective function could be more uniform and the back-propagation operations can stick on reaching the global optimum effectively, thus our deep multi-task learning algorithm can obtain more discriminative base deep CNNs and multi-task softmax for enhancing their separability significantly.
\begin{table}
\caption{\bf The comparisons on the accuracy rates (top 1) for 7 task groups: w/o MT means that multi-task learning is not performed; w MT means that multi-task learning is performed.}
\begin{center}
\begin{tabular}{| l  | c | c | c| c| c| c| c|}
\hline
Group No & 1 & 2 & 3 & 4 & 5 & 6 & 7\\
\hline 
 w/o MT(\%) & 42.3 & 37.9 & 30.7 & 32.3 & 43.5 & 48.0 & 39.3 \\
\hline 
 w MT(\%) & 43.6 & 38.8 & 31.9 & 33.7 & 44.2 & 49.3 & 40.4 \\					
\hline
\end{tabular}
\end{center}\label{tab:MTwo}
\end{table}
\begin{figure}[t]
\vspace*{-0.288cm}
\begin{center}
\includegraphics[width=0.418\textwidth]{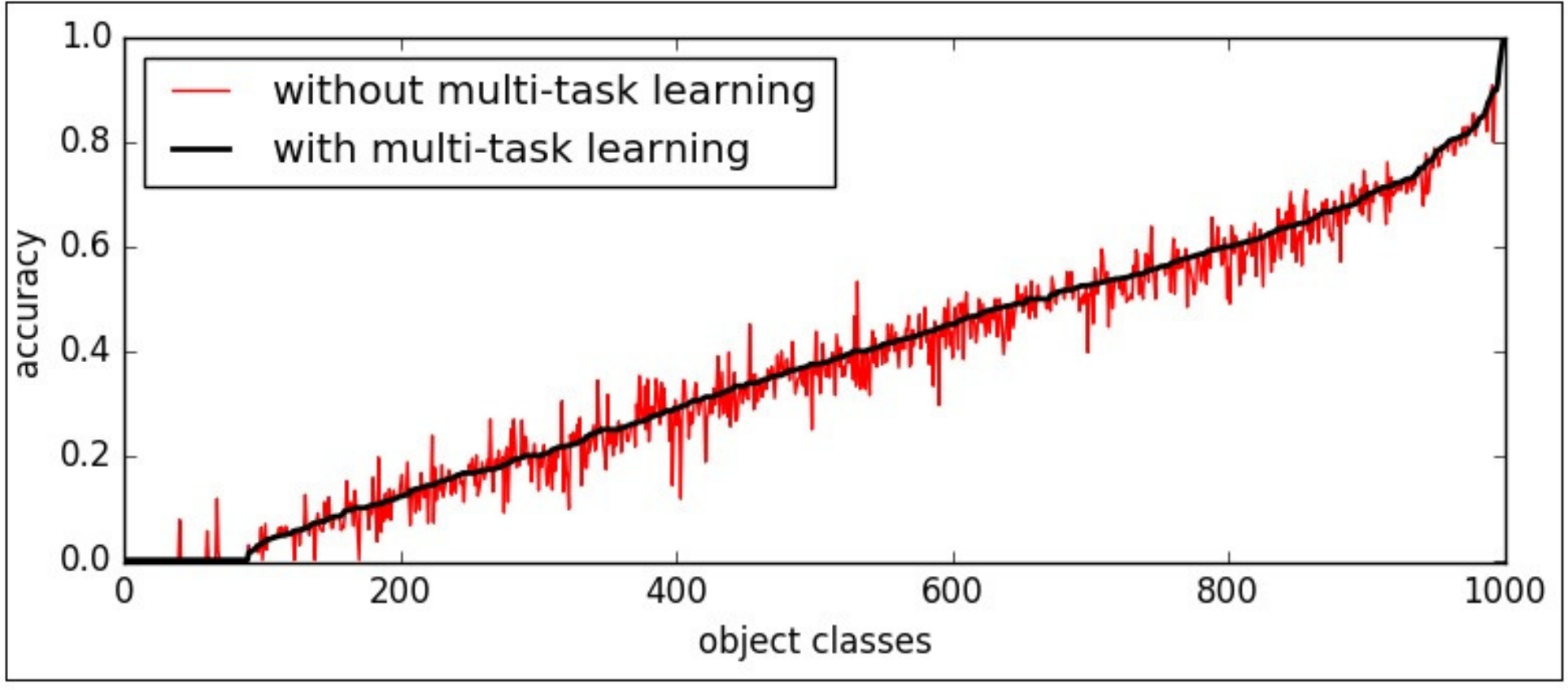}
\end{center}
\vspace*{-0.488cm}
   \caption{\bf The comparisons on the accuracy rates for 1,000 object classes in Task Group 1. }
\label{fig:long}
\label{fig:onecol}
\vspace*{-0.188cm}
\end{figure}
\begin{figure}[t]
\vspace*{-0.288cm}
\begin{center}
\includegraphics[width=0.418\textwidth]{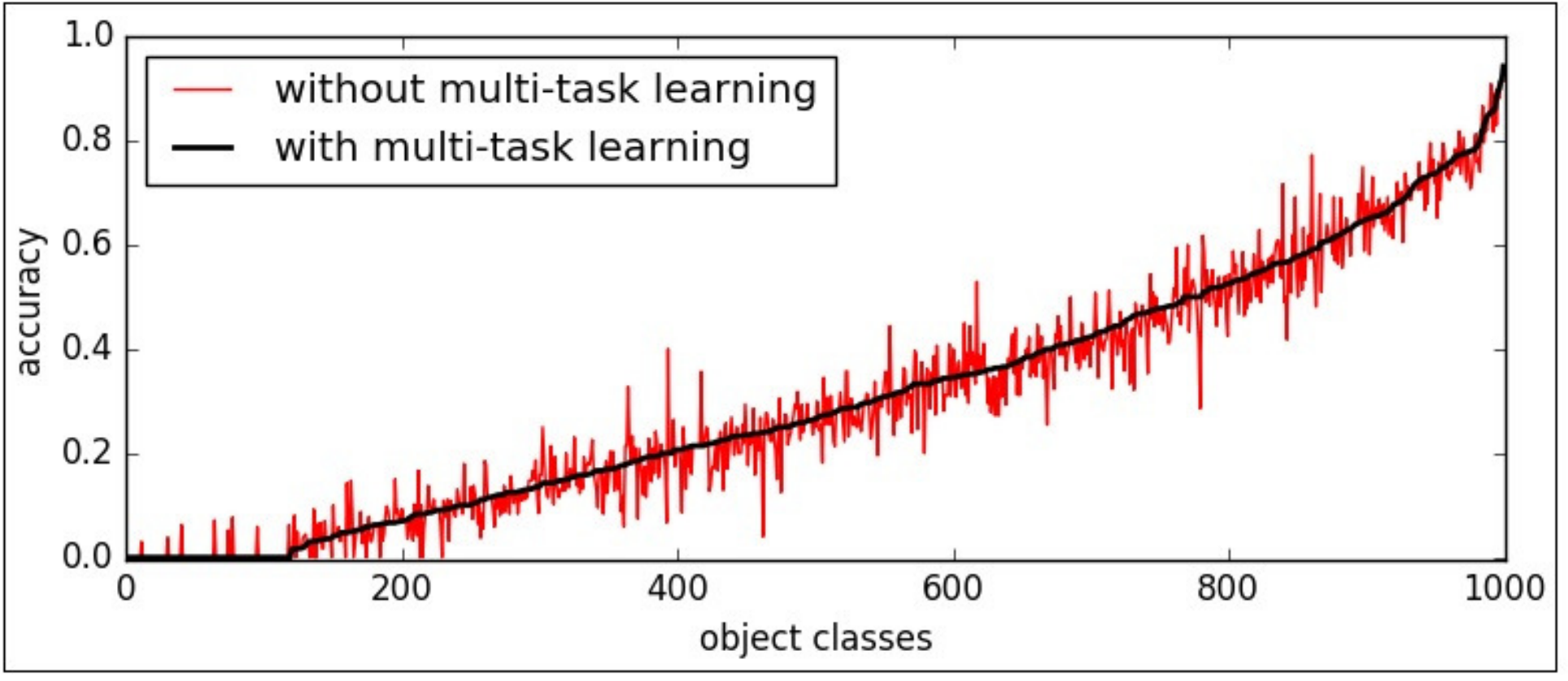}
\end{center}
\vspace*{-0.488cm}
   \caption{\bf The comparisons on the accuracy rates for 1,000 object classes in Task Group 5. }
\label{fig:long}
\label{fig:onecol}
\vspace*{-0.188cm}
\end{figure}
\begin{figure}[t]
\vspace*{-0.288cm}
\begin{center}
\includegraphics[width=0.418\textwidth]{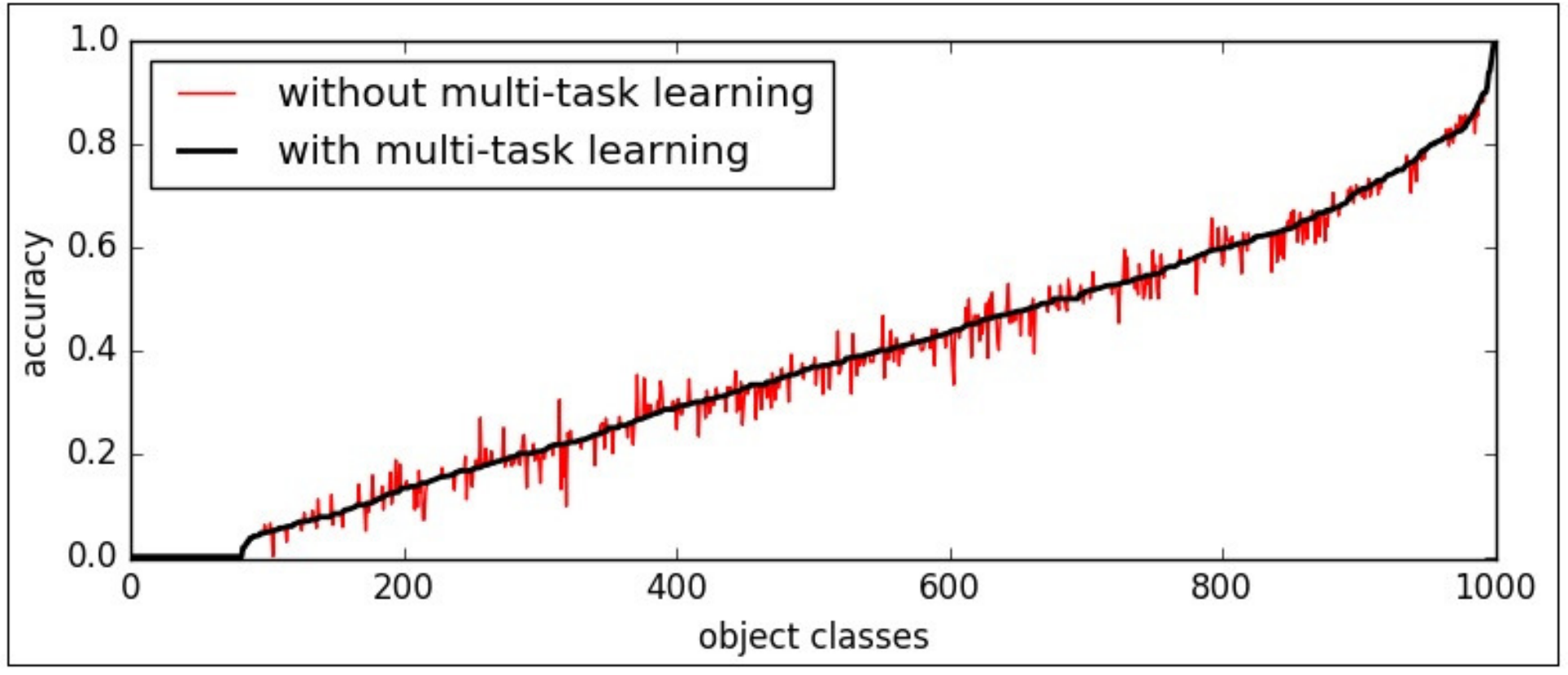}
\end{center}
\vspace*{-0.488cm}
   \caption{\bf The comparisons on the accuracy rates for 1,000 object classes in Task Group 7. }
\label{fig:long}
\label{fig:onecol}
\vspace*{-0.188cm}
\end{figure}

{\bf (c) Number of base deep CNNs $\vartheta$:}  By combining a set of base deep CNNs with diverse outputs (task spaces),  our deep mixture of diverse experts algorithm can achieve higher accuracy rates on recognizing $7,756$ atomic object classes in ImageNet10K image set. The reasons are three folds: (1) Each atomic object class can be assigned into multiple task groups, so that it can receive more sufficient comparison and contrasting with others from different aspects, which may significantly increase its chance to be distinguished from others; (2) The semantically-related atomic object classes are assigned into the same task group, and our deep multi-task learning algorithm can leverage their inter-class visual similarities to learn their base deep CNNs and multi-task softmax jointly to enhance their separability; (3) The atomic object classes in the same task group may share similar learning complexities, thus the gradients of their joint objective function could be more uniform and the back-propagation operations can stick on reaching the global optimum effectively.  Obviously, when more base deep CNNs are combined (i.e., allowing higher inter-group overlapping percentages), our deep mixture of diverse experts can achieve higher accuracy rates on large-scale visual recognition as shown in Fig. 13. 
\begin{figure}[t]
\vspace*{-0.288cm}
\begin{center}
\includegraphics[width=0.388\textwidth]{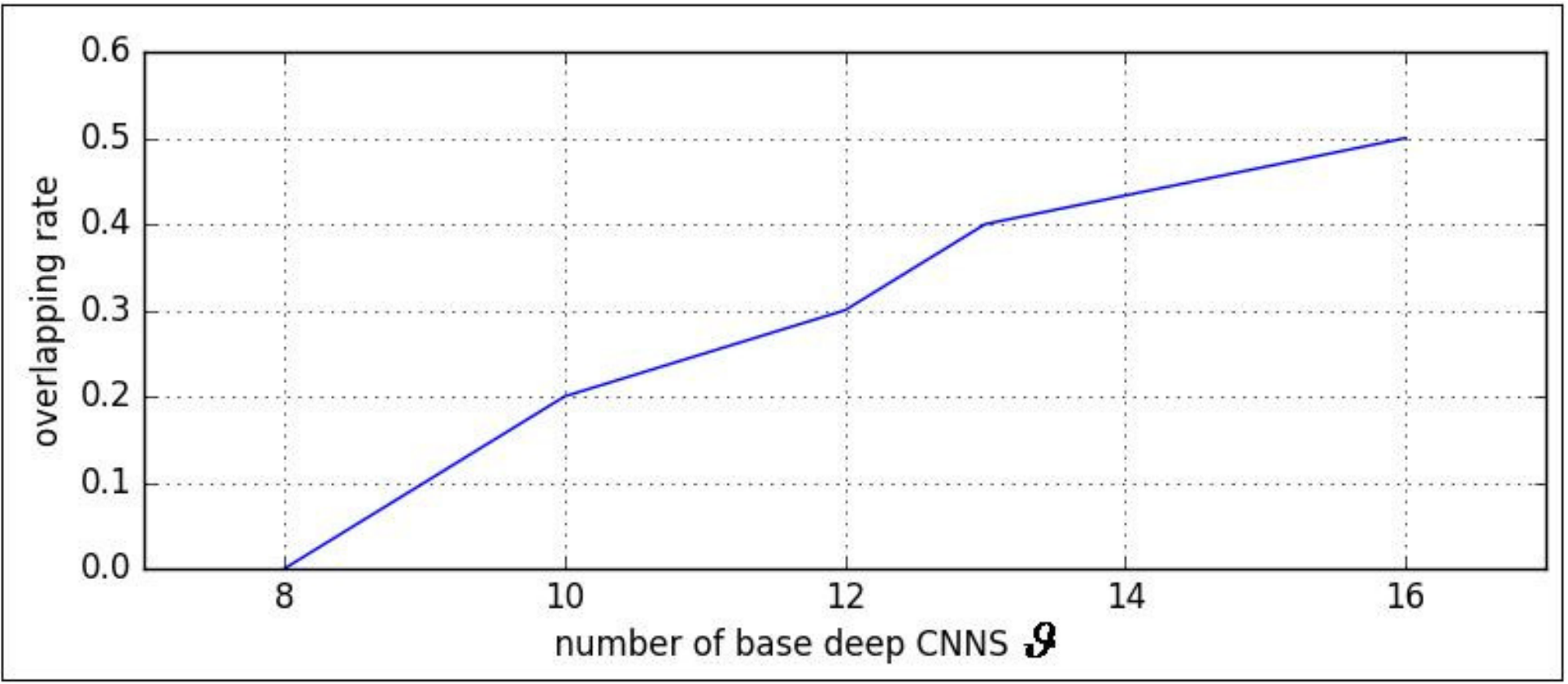}
\end{center}
\vspace*{-0.488cm}
   \caption{\bf The relationship between the average accuracy rates and the number of base deep CNNs being combined. }
\label{fig:long}
\label{fig:onecol}
\vspace*{-0.188cm}
\end{figure}

{\bf (d) Effectiveness of inter-group overlapping:} Different inter-group overlapping percentages can be used  for task group generation, so that each atomic object class can be assigned into different numbers of task groups (base deep CNNs). If each atomic object class can be assigned into more task groups and receive the predictions from more base deep CNNs, we can expect that it can have higher chance to be separated from others, e.g., increasing the inter-group overlapping percentages $\lambda$ may result in higher accuracy rates on large-scale visual recognition. As shown in Fig. 14, one can observe that our deep mixture of diverse experts algorithm can achieve better performance when more inter-group overlapping percentages $\lambda$ are used. The comparisons on the average accuracy rates are illustrated in Table III, where different inter-group overlapping percentages $\lambda$ are evaluated. 
\begin{figure}[t]
\vspace*{-0.288cm}
\begin{center}
\includegraphics[width=0.418\textwidth]{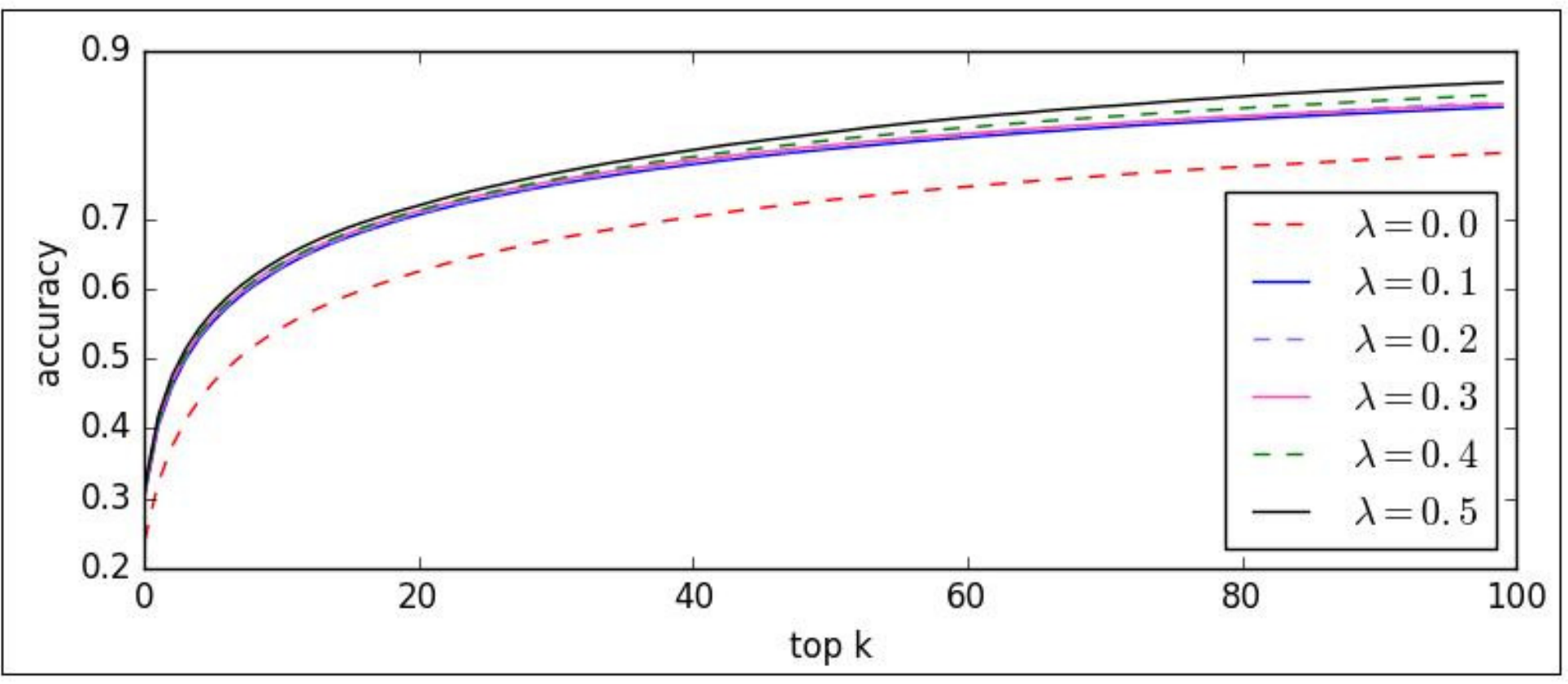}
\end{center}
\vspace*{-0.488cm}
   \caption{\bf The comparisons on the average accuracy rates when different inter-group overlapping percentages $\lambda$ are used for task group generation in our deep mixture of diverse experts algorithm.  }
\label{fig:long}
\label{fig:onecol}
\vspace*{-0.188cm}
\end{figure}

\begin{table}
\caption{\bf The performance comparisons when different inter-group overlapping percentages $\lambda$ are used. }
\begin{center}
\begin{tabular}{ |l|c|c|c|c|}
\hline
\multirow{2}{*}{$\lambda$} & \multicolumn{4}{|c|}{ accuracy rate (top k)} \\\cline{2-5}
& 1  &2 &5 &10  \\
\hline 
 
$\lambda$ = 0.0					& 33.96\% 	& 36.39\% 	& 48.17\% 	&57.31\%\\
\hline
$\lambda$ = 0.1					& 34.43\% 	& 37.54\% 	& 50.95\% 	&59.75\%\\
\hline
$\lambda$ = 0.2					& 35.66\% 	& 38.29\% 	& 51.17\% 	&60.99\%\\
\hline
$\lambda$ = 0.3					& 36.92\% 	& 39.03\% 	& 52.53\% 	&62.46\%\\
\hline
$\lambda$ = 0.4					& 37.95\% 	& 40.12\% 	& 54.38\% 	&63.21\%\\
\hline
$\lambda$ = 0.5					& \textbf{38.65\%} & \textbf{41.28\%}	 &\textbf{55.41\%} &\textbf{64.32\%}  \\
\hline
\end{tabular}
\end{center}
\end{table}
\begin{figure}[t]
\vspace*{-0.188cm}
\begin{center}
\includegraphics[width=0.48\textwidth]{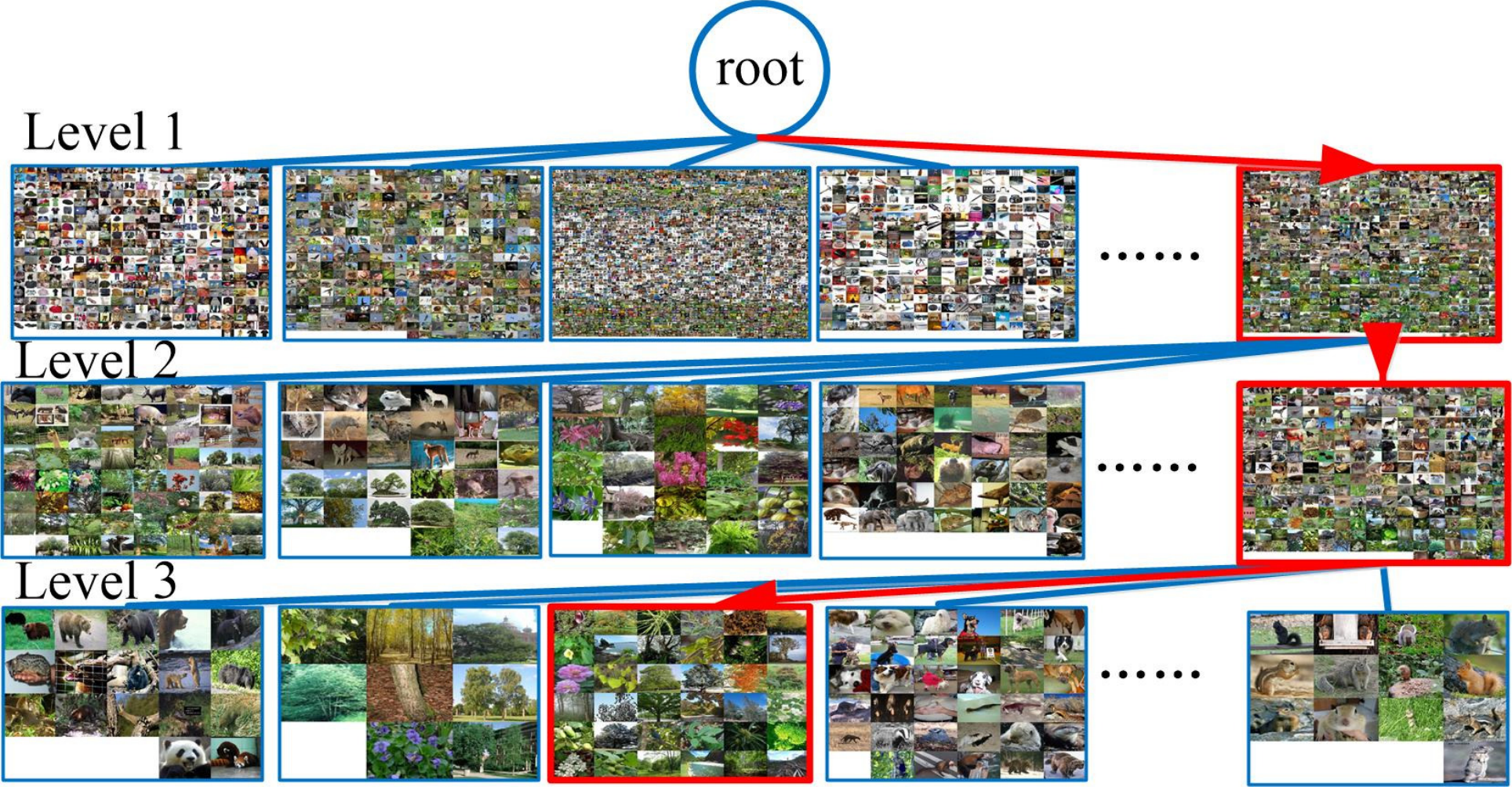}
\end{center}
\vspace*{-0.488cm}
   \caption{\bf Subtrees of our visual tree for organizing $7,756$ atomic object classes in ImageNet10K image set.}
\label{fig:long}
\label{fig:onecol}
\vspace*{-0.118cm}
\end{figure}

{\bf (e) Using different types of tree structures for task assignment:} Different tree structures can be used to organize large numbers of atomic object classes hierarchically: (1) our two-layer ontology; and (2) label tree and visual tree [28-35].  Our two-layer ontology can organize 7,756 atomic object classes hierarchically according to their inter-class semantic relationships, on the other hand, our visual tree can also be used to organize the same set of 7,756 atomic object classes hierarchically according to their inter-class visual similarities [34-35]. One experimental result on visual tree construction is shown in Fig. 15, where the pre-trained deep CNNs (AlexNet [11-13]) is used to extract the deep features for determining the inter-class visual similarities. It is very attractive to leverage these two tree structures for supporting task group generation in our deep mixture of diverse experts algorithm and compare their performances on large-scale visual recognition. As shown in Fig. 16 and Table I, one can observe that our deep mixture of diverse experts algorithm (when our two-layer ontology is used for task group generation) is slightly better than that when the visual tree is used for task group generation. The reasons for this phenomenon are: (1) The pre-trained deep CNNs (AlexNet [11-13] for 1,000 object classes) may be insufficient and ineffective to extract discriminative representations for 7,756 atomic object classes and learn their inter-class visual similarities accurately; (2) Using inaccurate inter-class visual similarities may not be able to learn the visual tree correctly; (2) Using the incorrect visual tree for task group generation may not be able to assign the visually-similar atomic object classes into the same task group, as a result, our deep mixture of diverse experts algorithm may obtain lower accuracy rates on large-scale visual recognition. 
\begin{figure}[t]
\vspace*{-0.188cm}
\begin{center}
\includegraphics[width=0.418\textwidth]{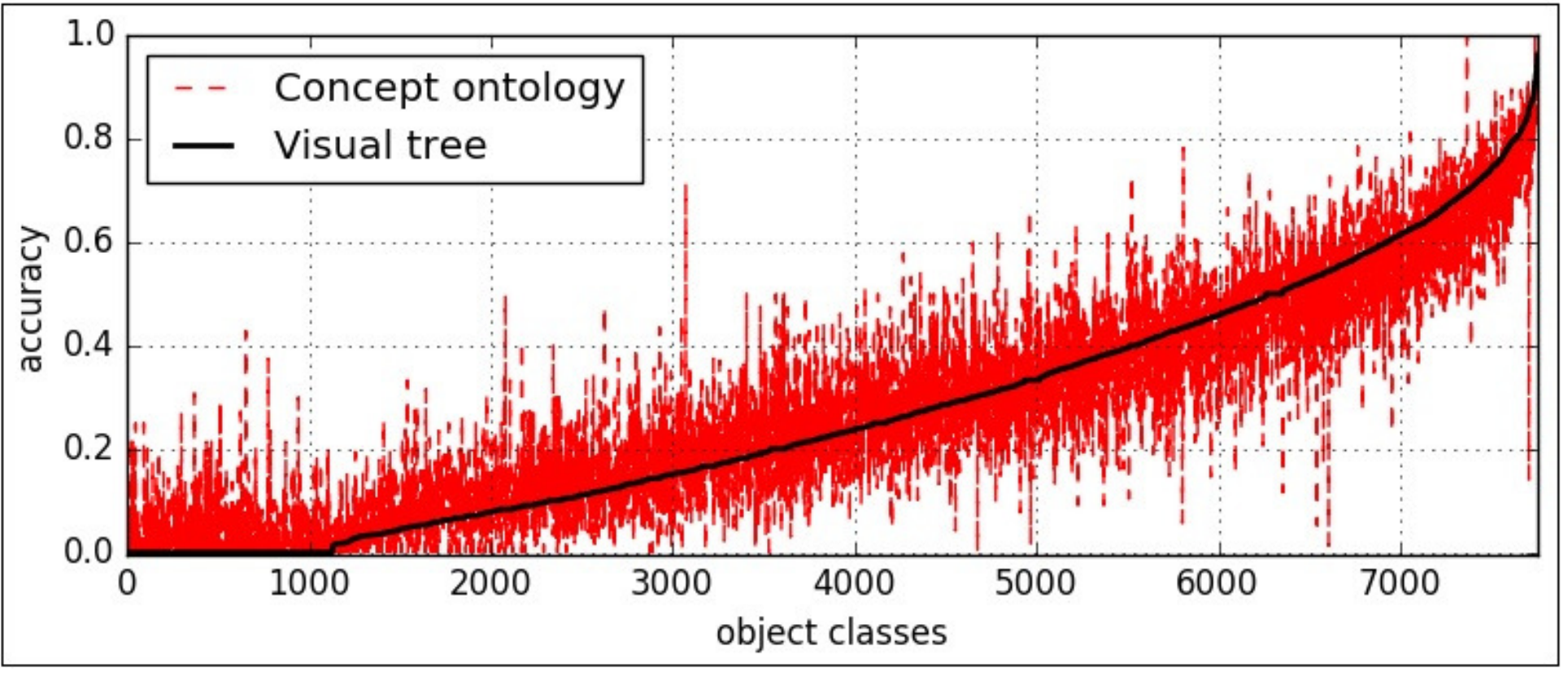}
\end{center}
\vspace*{-0.488cm}
   \caption{\bf The comparisons on the accuracy rates for 7,756 atomic object classes when different tree structures are used for task assignment: (a) two-layer ontology; (b) visual tree. }
\label{fig:long}
\label{fig:onecol}
\vspace*{-0.188cm}
\end{figure}

{\bf (f) Using different stacking functions:} It is worth noting that we can use different stacking functions to integrate the diverse outputs from all these base deep CNNs to generate 7,756-D high-level features for image semantics representation, thus it is very attractive to evaluate the effectiveness of our deep mixture of diverse experts algorithm when different stacking functions are used. In this paper, an alternative stacking function is defined, for a given image or object proposal, the $i$th component $\Upsilon(i)$ in the 7,756-D high-level features is used to indicate the appearance probability of the $i$th atomic object class $c_i$  and it is defined as: 
$$\Upsilon(i) =  \sum_{j=1}^{\vartheta} \lambda \Lambda_j (c_i) PS(i,j) \phi_j  \eqno(12)$$ 
where the indication function $\Lambda_j (c_i)$ is defined as: 
$$
\Lambda_j(c_i) = 
  \begin{cases} 
   1,	 & \text{if } c_i \text{ is in the jth task group }  \\
   \\
   0,      & \text{otherwise }
  \end{cases}
\eqno(13)$$

We have compared the performances of our deep mixture of diverse experts algorithm when different stacking functions (as defined in Eq. (8) and Eq. (12)) are used to integrate the diverse outputs from the same set of base deep CNNs to generate 7,756-D high-level features for image semantics representation, and the comparison results are shown in Fig. 17 and Table I (our deep mixture of diverse experts algorithm {\em vs.} Stack 2).  The first stacking function defined in Eq. (8) (our deep mixture of diverse experts algorithm) emphasizes the effect of the special class of ``not-in-group",  and it works better than the second stacking function defined in Eq. (12) (Stack 2). One can observe that two stacking functions have very similar performances, the reasons for this phenomenon are: (1) the predictions for the special class of ``not-in-group" can achieve the highest scores for all these irrelevant base deep CNNs, thus each base deep CNNs can make confident predictions effectively; (2) inter-group overlapping can allow these base deep CNNs to pass necessary messages effectively and make their predictions to be comparable directly.

{\bf (g) Hard object classes:} Large-scale visual recognition (e.g., recognizing tens of thousands of atomic object classes) is still a challenging issue. As shown in Figs. 6-12, one can observe that there still have some atomic object classes which are hard to be recognized correctly and their accuracy rates are very low. Thus it is very attractive to evaluate: (1) what are those hard object classes in ImageNet10K image set; (2) why such atomic object classes are hard to be recognized; and (3) what are the potential directions to be exploited for enhancing the separability of such hard object classes. 
\begin{figure}[t]
\vspace*{-0.288cm}
\begin{center}
\includegraphics[width=0.418\textwidth]{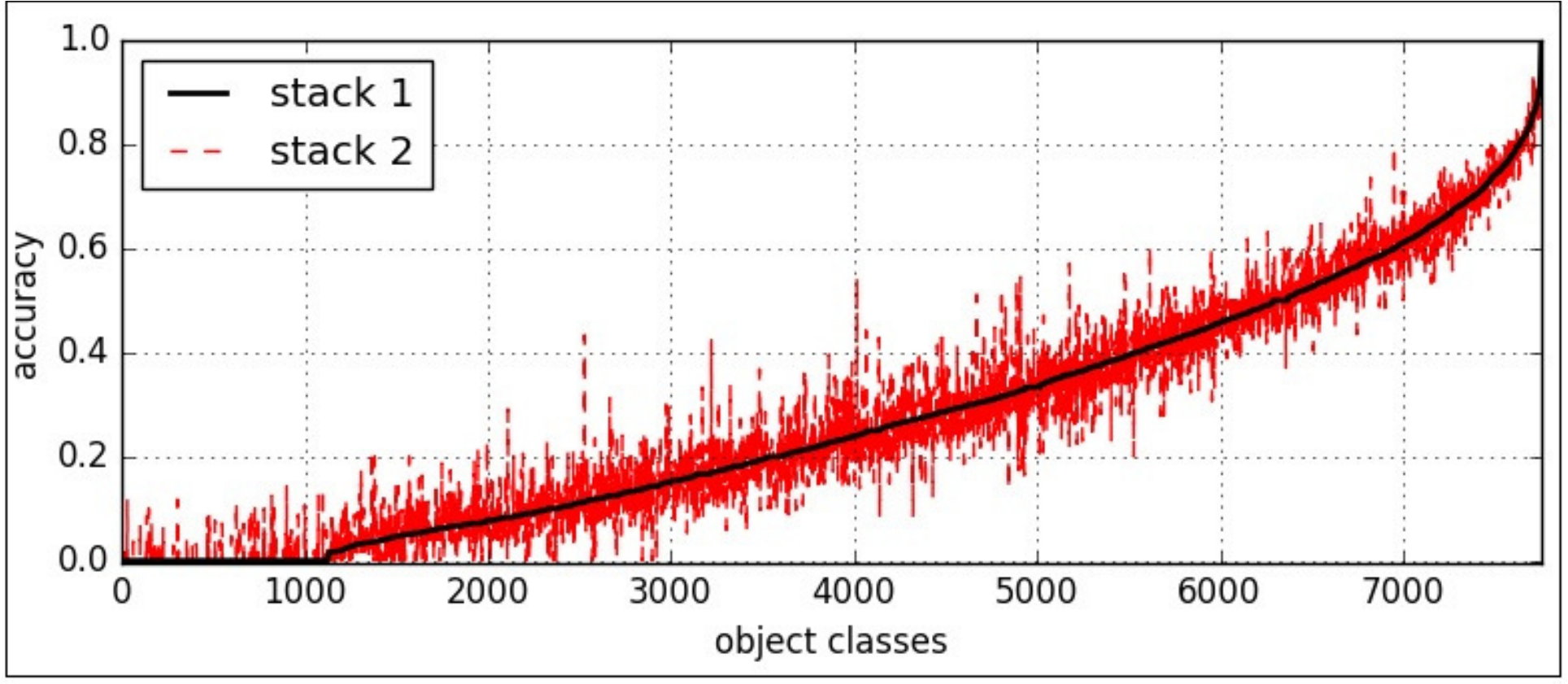}
\end{center}
\vspace*{-0.488cm}
   \caption{\bf The comparisons on the accuracy rates for 7,756 atomic object classes when different stacking functions are used to compose 7,756-D high-level features for image semantics representation. }
\label{fig:long}
\label{fig:onecol}
\vspace*{-0.188cm}
\end{figure}

The reasons for this phenomenon are: (1) Those hard object classes are fine-grained [7-10] and they are typically hard to be separated;  (2) The training images for those hard object classes have huge intra-class visual diversities and the available training images may be insufficient and inefficient to achieve accurate characterization of huge intra-class visual diversities;  (3) The ImageNet10K image set contains very rich visual contents and many object classes are unremarkable (i.e., they may not be able to be included in these 7,756 atomic object classes), but the appearances of such unremarkable object classes in the training images may mislead the joint process for learning the base deep CNNs and multi-task softmax. Thus it is very attractive to develop new algorithms that are robust to noisy samples and are able to achieve more sufficient and accurate characterization of huge intra-class visual diversities. 
\begin{table}
\caption{\bf The comparisons on the average accuracy rates.}
\begin{center}
\begin{tabular}{| l  | c | c| c|}
\hline
\multirow{1}{*}{approaches} & \multicolumn{3}{|c|}{accuracy rate (top k)} \\\cline{2-4}
& 1  & 5 & 10 \\
\hline 
 late fusion						& \textbf{38.65\%}	 &\textbf{55.41\%} &\textbf{64.32\%} 	\\
 \hline 
 early fusion & 36.23\%	& 52.45\% & 61.38\% 	\\ 

\hline
\end{tabular}
\end{center}\label{tab:10k}
\end{table}

{\bf (h) Early Fusion {\em vs.} Late Fusion:} 
In this paper, we combine a set of base deep CNNs at the softmax layer, e.g., late fusion by using the stacking function to integrate the diverse outputs from a set of base deep CNNs. Another alternative approach is early fusion, e.g., combining a set of base deep CNNs at the FC7 layer. In the alternative early fusion approach, the outputs of the FC7 layers (4,096-D features) from all these base deep CNNs are seamlessly integrated to generate 32,768-D features as the inputs of a 7,756-way softmax. This alternative early fusion approach is similar with bagging, each base deep CNNs focuses on one particular task group with $M$ ($M \leq 1,000$) atomic object classes, and all these 4,096-D features from a set of base deep CNNs are finally combined to generate 32,768-D features for image content representation and learn one 7.756-way softmax for large-scale visual recognition. As shown in Table IV, our late fusion approach can achieve better performance on large-scale visual recognition.  The reasons are: (1) our late fusion approach can effectively limit the misleading effects of the mistakes made by each base deep CNNs; (2) our stacking function can effectively leverage inter-group overlapping and message passing to make the predictions from all these base deep CNNs to be more comparable.

\section{Conclusions} 
Our deep mixture of diverse experts algorithm can seamlessly combine a set of base deep CNNs with diverse outputs (task spaces) to recognize tens of thousands of atomic object classes, where the structure of the well-designed AlexNet for 1,000 object classes is used to configure the base deep CNNs. By integrating a two-layer ontology to guide the process for task group generation, the semantically-related atomic object classes with similar learning complexities can be assigned into the same task group and our deep multi-task learning algorithm can leverage their inter-class visual similarities to learn more discriminative base deep CNNs and multi-task softmax jointly in an end-to-end fashion for enhancing their separability significantly. For the atomic object classes in the same task group, the gradients of their objective function are more uniform and the global optimum can be achieved effectively. Our deep mixture of diverse experts algorithm can build larger deep networks that are still cheap to compute at test time and more parallelizable at training time.  Our experimental results on ImageNet10K image set with $10,184$ image categories ($7,756$ atomic object classes) have demonstrated that our deep mixture of diverse experts algorithm can achieve very competitive results on large-scale visual recognition.

\end{document}